\documentclass{article}
\usepackage{microtype}
\usepackage{graphicx}
\usepackage{subcaption}
\usepackage{booktabs} 
\usepackage[table]{xcolor}
\usepackage{booktabs}
\usepackage{pifont}
\usepackage{multirow}
\usepackage{hyperref}

\usepackage{amsmath,amsfonts,bm}

\def\eqref#1{equation~\ref{#1}}

\def\1{\bm{1}}

\DeclareMathAlphabet{\mathsfit}{\encodingdefault}{\sfdefault}{m}{sl}
\SetMathAlphabet{\mathsfit}{bold}{\encodingdefault}{\sfdefault}{bx}{n}

\newcommand{\cmark}{\ding{51}} 
\newcommand{\xmark}{\ding{55}} 
\usepackage[accepted]{icml2026}

\usepackage{amsmath}
\usepackage{amssymb}
\usepackage{mathtools}
\usepackage{amsthm}
\usepackage[capitalize,noabbrev]{cleveref}
\usepackage{tcolorbox}
\theoremstyle{plain}

\theoremstyle{definition}

\theoremstyle{remark}

\usepackage{tabularx}

\usepackage[textsize=tiny]{todonotes}

\icmltitlerunning{Krause Synchronization Transformers}

\begin{document}

\twocolumn[
  \icmltitle{Krause Synchronization Transformers}



  \icmlsetsymbol{equal}{*}

  \begin{icmlauthorlist}
    \icmlauthor{Jingkun Liu}{qz,cai,sj}
    \icmlauthor{Yisong Yue}{caltech}
    \icmlauthor{Max Welling}{uva,cuspai}
    \icmlauthor{Yue Song}{qz,cai}
  \end{icmlauthorlist}

  \icmlaffiliation{qz}{Shanghai Qi Zhi Institute}
  \icmlaffiliation{cai}{College of AI, Tsinghua University}
  \icmlaffiliation{sj}{Shanghai Jiao Tong University}
  \icmlaffiliation{caltech}{California Institute of Technology}
  \icmlaffiliation{uva}{University of Amsterdam}
  \icmlaffiliation{cuspai}{Cusp AI}

  \icmlcorrespondingauthor{Yue Song}{songyue19960927@gmail.com}

  \icmlkeywords{Machine Learning, ICML}

  \vskip 0.3in
]



\printAffiliationsAndNotice{}  

\begin{abstract}

Self-attention in Transformers relies on globally normalized softmax weights, causing all tokens to compete for influence at every layer. When composed across depth, this interaction pattern induces strong synchronization dynamics that favor convergence toward a dominant mode, a behavior associated with representation collapse and attention sink phenomena. We introduce \textbf{Krause Attention}, a principled attention mechanism inspired by bounded-confidence consensus dynamics. Krause Attention replaces similarity-based global aggregation with distance-based, localized, and selectively sparse interactions, promoting structured local synchronization instead of global mixing. We relate this behavior to recent theory modeling Transformer dynamics as interacting particle systems, and show how bounded-confidence interactions naturally moderate attention concentration and alleviate attention sinks. Restricting interactions to local neighborhoods also reduces runtime complexity from quadratic to linear in sequence length. Empirically, we validate Krause Attention across diverse settings, including vision (ViT on CIFAR/ImageNet), autoregressive image generation (MNIST/CIFAR-10), large language models (Llama/Qwen), and language models trained from scratch at multiple scales (100M/200M). Across these domains, Krause Attention achieves consistent performance gains while improving computational efficiency, highlighting bounded-confidence dynamics as a scalable and effective inductive bias for attention. Project page is available at \url{https://jingkun-liu.github.io/krause-sync-transformers/}.

\end{abstract}


\begin{figure}[h]
    \centering
    \includegraphics[width=.99\linewidth]{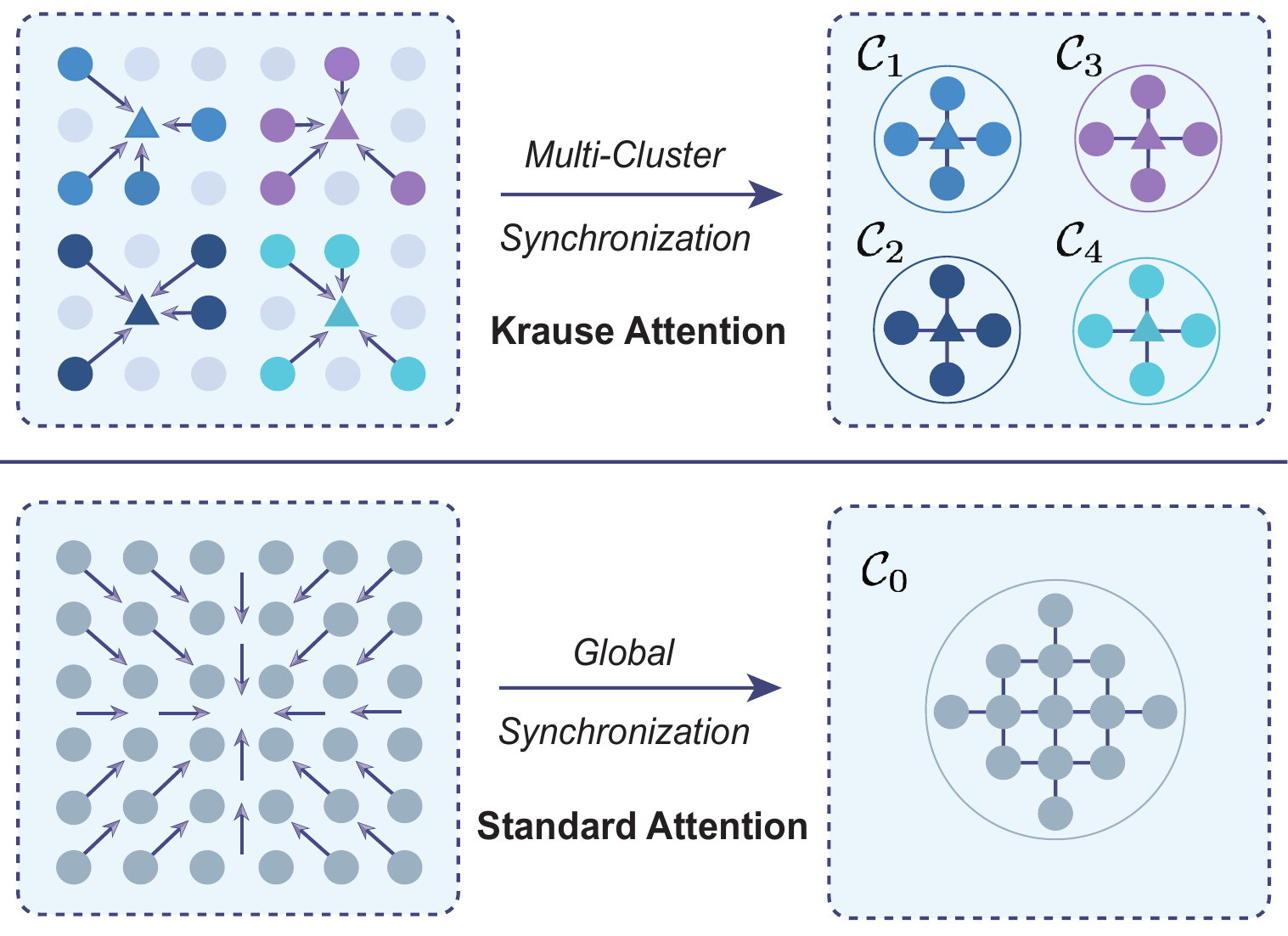}
    \caption{Krause Attention, grounded in bounded-confidence interactions, promotes localized multi-cluster synchronization (\emph{top}). In contrast, standard self-attention tends to induce globally coupled dynamics that concentrate attention onto a dominant mode, often manifesting as attention sinks~\cite{xiao2023efficient} (\emph{bottom}).}
    \label{fig:teaser}
\end{figure}

\section{Introduction}

Transformers~\citep{vaswani2017attention} have become the \emph{de facto} architecture across a wide range of domains, including vision, language, and generative modeling. Their success is primarily attributed to the self-attention mechanism, which enables flexible, content-dependent interactions among tokens. 
Despite their immense success, it remains an open question to deeply understand why self-attention is so effective, and how its design can be systematically improved.


Recent theoretical work characterizes self-attention dynamics as interacting particle systems, such as how tokens as particles evolve through the layers~\citep{geshkovski2023emergence}. Global normalization induces globally coupled dynamics that exhibit strong synchronization tendencies: in the mean-field limit, token representations contract toward a single dominant mode~\citep{chen2025quantitative}. This dominant mode can be connected to \emph{attention sink} phenomena, where attention mass collapses onto few tokens~\citep{xiao2023efficient}. Together, these findings suggest that globally normalized attention can induce undesirable coordination effects when iterated across depth, motivating a fundamental question:


\begin{tcolorbox}[
    colback=gray!10!white, 
    colframe=gray!70!black, 
    sharp corners=all 
]
\emph{Should globally normalized similarity govern token interactions by default?}
\end{tcolorbox}

To tackle this question, we propose a different principle: instead of global similarity-based aggregation, we explicitly model token interactions through \emph{bounded-confidence synchronization dynamics}. Our approach draws inspiration from the Krause consensus model~\citep{krause2000discrete,hegselmann2002opinion}, where interacting agents only influence neighbors within a bounded confidence radius, leading to stable multi-cluster structures rather than global consensus. We introduce \textbf{Krause Attention} (Fig.~\ref{fig:teaser}) as a Transformer-compatible instantiation of this idea. Krause Attention replaces query–key dot-product similarity with query–key \emph{distance}, which is mapped through a radial basis function (RBF) kernel to produce smooth, well-conditioned interaction weights. These distance-based affinities are further constrained by a local neighborhood and selective top-$k$ sparsity, enforcing a bounded-confidence interaction pattern in which tokens primarily interact only with nearby representations, while distant tokens exert little or no influence.

This design has direct implications for the resulting dynamics. Theoretically, we connect Krause Attention to recent analyses of Transformer clustering dynamics~\citep{geshkovski2023emergence,chen2025quantitative}. While standard self-attention induces globally coupled dynamics that favor convergence toward a single consensus, the bounded-confidence structure of Krause Attention supports \emph{stable multi-cluster formations}. This naturally moderates attention concentration and mitigates attention sink behavior. Locality further reduces computational cost from $\mathcal{O}(N^2d)$ to $\mathcal{O}(NWd)$, where $W$ is the local window size, and $d$ the token dimension.

Empirically, Krause Attention delivers consistent and substantial gains across vision, generation, and language modeling tasks.
For image classification, Krause Vision Transformers (ViTs) consistently outperform standard ViTs~\citep{dosovitskiy2021image} on CIFAR-10/100 and ImageNet-1K, achieving an \textbf{average accuracy improvement of $\mathbf{+3.0\%}$} while \textbf{reducing FLOPs by approximately $\mathbf{30\%}$} across model scales.
In autoregressive image generation~\citep{parmar2018image}, Krause-based models achieve \textbf{lower negative log-likelihood} than standard Transformers while enabling \textbf{more than $\mathbf{2\times}$ faster inference}. For large language models~\citep{team2024qwen2,grattafiori2024llama} and models trained from scratch, incorporating Krause Attention as an auxiliary pathway consistently improves \textbf{zero-shot performance over both LoRA-finetuned}~\citep{hu2022lora} \textbf{and from-scratch baselines} across a broad suite of language understanding benchmarks. 

Together, these results demonstrate that bounded-confidence dynamics provide a \textbf{scalable, computationally efficient, and practically effective inductive bias} for self-attention mechanisms across diverse modalities and model regimes.

\section{Related Work}

\noindent\textbf{Transformer Architectures.} Since their introduction, Transformers~\citep{vaswani2017attention} have become the prevailing architectural paradigms across a wide range of machine learning domains. Their success stems from the scalable self-attention mechanisms and flexible token representations, which have enabled consistent performance improvements with increasing model and data size. In visual recognition, Vision Transformers~\citep{dosovitskiy2021image} and subsequent variants~\citep{liu2021swin,touvron2021training,rao2021dynamicvit,yuan2021tokens,han2021transformer,wang2021pyramid,touvron2022deit,heo2024rotary} have demonstrated that Transformer-based models can match or surpass convolutional networks when trained under comparable large-scale settings. Beyond discriminative tasks, Transformer-based architectures also underpin a broad spectrum of generative modeling frameworks, including autoregressive models~\citep{parmar2018image,chen2020generative,yu2022scaling,cao2021image,touvron2021going} and diffusion models built on Transformer backbones~\citep{he2022masked,chang2022maskgit,peebles2023scalable,wang2025lavin}. In language modeling, decoder-only Transformers scaled to hundreds of billions of parameters form the foundation of modern LLMs~\citep{achiam2023gpt,grattafiori2024llama,team2024qwen2,liu2025deepseek,comanici2025gemini}, further reinforcing the central role of Transformers.



A large body of work seeks to reduce the quadratic cost of standard self-attention. Sparse attention architectures \citep{child2019generating} restrict interactions to predefined patterns, while low-rank or kernel-based approaches such as Linformer \citep{wang2020linformer}, Performer \citep{choromanski2021rethinking}, and other linear-attention variants \citep{katharopoulos2020transformers} approximate attention through low-dimensional projections or random feature kernels. Hashing-based methods like Reformer~\citep{kitaev2020reformer} further achieve sub-quadratic complexity by clustering tokens. While these approaches improve scalability, they often weaken fine-grained, content-dependent interactions. Our work is orthogonal to this efficiency-oriented line of research. Instead of approximating softmax attention or imposing fixed sparsity patterns, we redesign the interaction rule itself by drawing on Krause bounded-confidence dynamics. Krause Attention introduces distance-based, content-adaptive sparsity that promotes localized multi-cluster structures. Although the use of local neighborhoods reduces the computational cost to $\mathcal{O}(N Wd)$, our primary goal is to endow attention with an explicit and theoretically grounded synchronization prior, which is distinct from prior sub-quadratic approximations.

\noindent\textbf{Understanding Transformers.} Substantial progress has been made toward interpreting Transformers through physical and mathematical lenses~\citep{bruno2025multiscale,ren2025transformers,rigollet2025mean}. Self-attention mechanisms have been connected to kernel methods~\citep{tsai2019transformer}, continuous Hopfield networks~\citep{ramsauer2021hopfield}, and energy-based associative memories~\citep{yang2022transformers,hoover2023energy,hu2024provably}. Complementary studies have examined relationships with convolution operators~\citep{cordonnier2020relationship} and non-convex optimization landscapes~\citep{hu2024provably,abbe2024far}. From a dynamical system perspective, the evolution of token representations is often modeled as an interacting particle system~\citep{geshkovski2023emergence}, with links to global synchronization phenomena such as Kuramoto models~\citep{kuramoto2005self,acebron2005kuramoto}. Recent analyses further establish quantitative bounds on convergence toward mean-field limits~\citep{chen2025quantitative,rigollet2025mean,bruno2025emergence}. Inspired by~\citet{geshkovski2023emergence}, we adopt Krause opinion dynamics \citep{krause2000discrete,hegselmann2002opinion} as a principled mechanism for structuring attention. Incorporating bounded confidence into the attention kernel enforces localized, query-key distance-aware token interactions rather than global mixing. As a result, the induced dynamics align with theoretical analyses showing that restricting global coupling can instead support the emergence of multiple coherent clusters.

\begin{figure*}[t]
    \centering
    \includegraphics[width=0.87\linewidth]{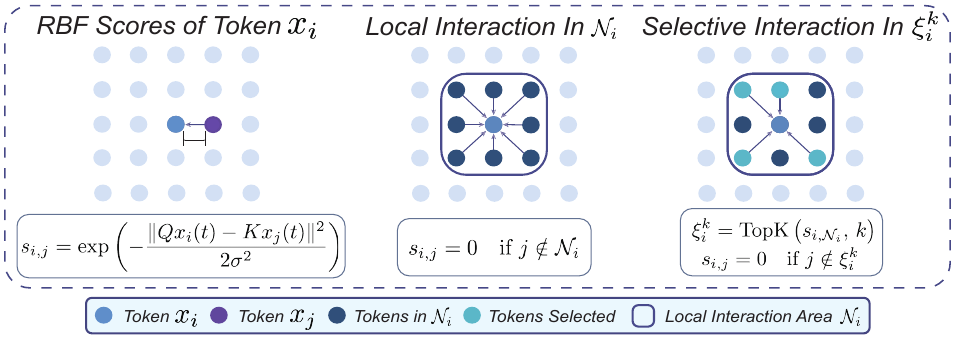}
    \caption{Krause Attention computes RBF affinity scores, restricts updates to spatial local neighborhoods, and applies top-$k$ selective interactions in the representation space.}
    \label{fig:Krause Attention mechanism}
\end{figure*}

\section{Preliminaries}

\subsection{Self-attention Mechanism}

Self-attention is the central component underlying the Transformer architecture. Given an input sequence $X \in \mathbb{R}^{N \times d}$, where $N$ denotes the sequence length and $d$ the embedding dimension, the input is linearly projected into queries, keys, and values through learnable weight matrices:
\begin{equation}
Q = XW_Q, \quad K = XW_K, \quad V = XW_V,
\end{equation}
where $Q, K \in \mathbb{R}^{N \times d_k}$ and $V \in \mathbb{R}^{N \times d_v}$ denote the resulting query, key, and value representations. The standard scaled dot-product attention computes the output for each token as a weighted aggregation of all value vectors, where the weights depend on the similarity between queries and keys. Specifically, let $\mathbf{q}_i, \mathbf{k}_j \in \mathbb{R}^{1 \times d_k}$ and $\mathbf{v}_j \in \mathbb{R}^{1 \times d_v}$ denote the $i$-th (or $j$-th) row vectors of the weight matrices. The output representation for token $i$ is given by:
\begin{equation}
\mathbf{z}_i = \sum_{j=1}^{N} \text{Softmax} \left( \frac{\mathbf{q}_i \mathbf{k}_j^T}{\sqrt{d_k}} \right) \cdot \mathbf{v}_j.
\end{equation}
Stacking all $\mathbf{z}_i$ yields the output matrix $Z \in \mathbb{R}^{N \times d_v}$. Through dot-product similarity and global softmax normalization, this formulation defines a fully coupled interaction rule in which each token aggregates information from all others. This globally normalized coupling implicitly hard-codes a particular coordination dynamic whose long-term behavior is neither explicitly modeled nor controlled, and may induce undesirable collective effects. This motivates alternative principles that explicitly govern token interaction.




\subsection{Krause Consensus Model}


The Krause consensus model describes a non-linear, discrete-time dynamical system over a population of $N$ agents, where interaction strengths depend on the agents’ current states. Let $x(t) \in \mathbb{R}^N$ denotes the vector of agent states at iteration $t$. The system evolves according to:
\begin{equation}
x(t+1)=A\big(x(t),t\big)\ x(t),
\end{equation}
where $A\big(x(t),t\big)\in\mathbb{R}^{N\times N}$ is a row-stochastic influence matrix whose entries $a_{i, j}\big(x(t),t\big)$ quantify the influence of agent $j$ on agent $i$. A defining feature of the model is the \emph{bounded confidence} mechanism, which restricts interactions to agents whose states lie within a confidence radius $\epsilon$. For each agent $i$, the active neighbor set is defined as
\begin{equation}
I\big(i,x(t)\big) = \left\{ j \in {1,\dots,N} \mid |x_i(t) - x_j(t)| \le \epsilon \right\}.
\end{equation}
The influence weights are assigned uniformly within this active neighborhood, with $a_{i, j}\big(x(t),t\big) = |I\big(i,x(t)\big)|^{-1}$ if $j \in I\big(i,x(t)\big)$ and $0$ otherwise.

This state-dependent nonlinearity drives the system toward a partitioned equilibrium. In the limit, agents self-organize into disjoint, invariant clusters, within which local consensus emerges while distinct clusters remain separated. 

Our Krause Attention mechanism instantiates the Krause consensus model within the Transformer architecture by treating tokens as agents and their value embeddings as agent states.  Under this mapping, attention weights define state-dependent influence matrices that govern how token representations are updated, mirroring the bounded-confidence interactions of the classical Krause model. 








\section{Krause Attention}

This section introduces Krause Attention (see Fig.~\ref{fig:Krause Attention mechanism}), the core mechanism of Krause Synchronization Transformers. Secs.~\ref{sec:4.1_QKD} and~\ref{sec:4.2_CVLM} detail the key components of the method, while Sec.~\ref{sec:4.3_CTCD} connects Krause Attention to clustering and synchronization analyses of Transformer models.


\subsection{Distance-Aware Query-Key Interactions}
\label{sec:4.1_QKD}

To translate the Krause consensus model into an attention mechanism, a central requirement is a principled notion of similarity between Transformer tokens, which are interpreted as interacting agents. In our framework, token interactions are governed by a distance-based measure of opinion similarity. Instead of the dot-product similarity used in standard Transformers, we explicitly quantify dissimilarity using the Euclidean distance between query and key representations. Formally, given query vectors ${q}_i$ and key vectors ${k}_j$, we define the pairwise query–key distance as
\begin{equation}
\label{distance}
\Delta_{i, j} = ||{q}_i - {k}_j|| = \sqrt{\sum_{n=1}^{d_k} (q_{i,n} - k_{j,n})^2},
\end{equation}
which serves as a measure of opinion dissimilarity between tokens $i$ and $j$. To map distances into interaction strengths, we apply a radial basis function (RBF) kernel to the squared distances, yielding unnormalized affinity scores
\begin{equation}
\label{sij_formulation}
s_{i, j} = \exp(- \frac{\Delta_{i, j}^2}{2\sigma^2}),
\end{equation}
where $\sigma$ is a learnable scale parameter which controls the effective interaction range. This kernel assigns higher affinity to tokens with similar query–key representations, while naturally suppressing interactions between distant tokens. Importantly, the RBF distance admits a \textbf{\emph{separable}} formulation with respect to queries and keys. This distance-based structure induces localized, distance-aware interactions and naturally facilitates structured sparsity (see Sec.~\ref{sec:4.2_CVLM}).


The final dense attention weights are obtained by normalizing these affinities across the sequence dimension:
\begin{equation}
\label{eq:dense_attention}
a_{i,j} = \frac{s_{i,j}}{\sum_{n=1}^{N} s_{i,n}},
\end{equation}
which ensures a valid probability distribution over tokens and induces global interactions across all tokens.

Notably, unlike standard attention, we do not apply an additional \texttt{softmax} operation to the RBF scores. The exponential nonlinearity and temperature-like scaling introduced by \texttt{softmax} are already captured by the RBF kernel through its exponential form and variance parameter $\sigma^2$. As a result, this formulation produces distance-aware, smoothly decaying attention weights that naturally align with the bounded-confidence interactions characteristic of Krause dynamics.


\subsection{Local and Selective Interactions}
\label{sec:4.2_CVLM}

The dense attention in Eq.~(\ref{eq:dense_attention}) enables unrestricted global information exchange, allowing every token to interact with all others. While effective in standard Transformers, such fully coupled mixing induces strong global competition and repeated averaging, which can obscure local structure and amplify dominant tokens across layers. To address this, we introduce \textbf{\emph{local}} and \textbf{\emph{selective}} interaction constraints, which are central to bounded-confidence consensus models.

\noindent\textbf{Local Interactions.} We introduce a local interaction mechanism that restricts attention to a pre-defined neighborhood $\mathcal{N}_i$ for each token $i$. Instead of normalizing attention weights over the full sequence, the resultant normalization is performed only within this neighborhood:
\begin{equation}
\label{local_mixing}
\tilde{a}_{i,j}
= \frac{s_{i,j}}{\sum_{\ell \in \mathcal{N}_i}s_{i,\ell}},
\qquad j \in \mathcal{N}_i,
\end{equation}
so that token $i$ aggregates information exclusively from locally admissible neighbors. This explicit locality constraint enforces a \emph{hard interaction cutoff} analogous to bounded confidence in Krause dynamics, suppressing global averaging while preserving meaningful local interactions.

The neighborhood $\mathcal{N}_i$ is defined according to the underlying task structure. In vision models, tokens are associated with spatial locations, and $\mathcal{N}_i$ corresponds to a local spatial window centered at token $i$. In autoregressive settings, including image generation and language modeling, $\mathcal{N}_i$ is instead specified as a causal window that limits attention to a fixed number of preceding tokens.

\noindent\textbf{Selective Interactions.} While local interaction enforces spatial or temporal locality, bounded-confidence dynamics further imply that influence should be both limited and competitive, with only a small subset of neighbors exerting meaningful impact. To operationalize this principle, we introduce a mechanism that sparsifies attention within each local neighborhood. Specifically, for each token $i$, we select the top-$k$ neighbors in $\mathcal{N}_i$ according to their similarity scores. Let $\xi_i^k \subseteq \mathcal{N}_i, |\xi_i^k|=k$ denote the set of indices corresponding to the $k$ largest values of $s_{i,j}$ for $j\in\mathcal{N}_i$. Attention weights are then normalized over this reduced set:
\begin{equation}
\label{final_weight}
\boxed{\tilde{a}^*_{i,j}
= \frac{s_{i,j}}
{\sum_{\ell \in \xi_i^k} s_{i,\ell}}, \qquad j \in \xi_i^k,}
\end{equation}
ensuring that token $i$ interacts only with its most relevant local neighbors. \textbf{This top-$k$ formulation constitutes the final attention rule used in Krause Attention.} The output representation is then computed as $z_i = \sum_{j \in \mathcal{N}_i} \tilde{a}^*_{i,j}\cdot v_j$.

This selective interaction mechanism induces a functionally sparse-attention pattern, in which each token aggregates information from a small, adaptively chosen subset of neighbors. By combining local interaction with the selective top-$k$ selection, our Krause Attention departs from the dense, fully connected interaction graph of standard self-attention and instead recovers the characteristic \textbf{\emph{local clustering with global coherence}} behavior of Krause consensus dynamics.

\noindent\textbf{Time Complexity and Training Algorithm.} Krause Attention restricts interactions to a local neighborhood and further prunes them via top-$k$ selective sparsity. As a result, each token attends only to $W$ nearby candidates and retains $k \le W$ of them during selective mixing. This yields an overall time complexity $\mathcal{O}(NWd)$, which is strictly lower than the $\mathcal{O}(N^2)$ cost of self-attention while preserving content-adaptive token interactions. Algorithm \ref{alg:krause} summarizes the forward pass of a Krause Attention layer. The procedure follows three steps: (1) compute query–key distances and transform them using an RBF kernel, (2) apply locality masking followed by top-$k$ selection within the local region, and (3) normalize affinities and aggregate values.





\begin{algorithm}[h]
\caption{Forward pass of a Krause Attention layer}
\label{alg:krause}
\begin{algorithmic}[1]
\STATE \textbf{Input:} token sequence $x$, local neighborhoods $\{\mathcal{N}_i\}$, sparsity level $k$, kernel scales $\{\sigma_h\}$
\STATE \textbf{Output:} updated token sequence $z$

\STATE Compute projections: $(q,k,v) \leftarrow \texttt{Linear}(x)$

\FOR{each head $h = 1,\dots,H$}

    \STATE \textbf{(1) Distance-based affinity}
    \STATE $\Delta_{i,j} \leftarrow \| q_i - k_j \|^2$ \hfill Eq.~(\ref{distance})
    \STATE $s_{i,j} \leftarrow \exp\!\left(-\Delta_{i,j}/(2\sigma_h^2)\right)$ \hfill Eq.~(\ref{sij_formulation})

    \STATE \textbf{(2) Apply locality}
    \STATE $s_{i,j} \leftarrow 0 \quad \text{if } j \notin \mathcal{N}_i$ \hfill Eq.~(\ref{local_mixing})

    \STATE \textbf{(3) Selective top-$k$ sparsity}
    \STATE $\xi_i^k \leftarrow \operatorname{TopK}\big(s_{i,\mathcal{N}_i},\, k\big)$
    \STATE $s_{i,j} \leftarrow 0 \quad \text{if } j \notin \xi_i^k$ 

    \STATE \textbf{(4) Normalization and aggregation}
    \STATE $a_{i,j} \leftarrow s_{i,j} \Big/ \sum_{\ell \in \xi_i^k} s_{i,\ell}$ \hfill Eq.~(\ref{final_weight})
    \STATE $o_h[i] \leftarrow \sum_{j \in \xi_i^k} a_{i,j} \, v_j$

\ENDFOR

\STATE $z \leftarrow \texttt{Linear}_{out}(\texttt{Concat}(o_1,\dots,o_H))$

\STATE \textbf{return} $z$
\end{algorithmic}
\end{algorithm}

\subsection{Connection to Transformer Clustering Dynamics}
\label{sec:4.3_CTCD}

This subsection situates Krause Attention within recent theoretical analyses of Transformers' clustering behavior. Within this perspective, Krause Attention can be understood as a modification of standard self-attention flows, converting the fragile, architecture-dependent clustering behavior into a more explicit and controllable mechanism.


\noindent\textbf{Background: Clustering in Self-attention.}
Recent work has established that self-attention layers can be interpreted as dynamical systems acting on token representations. In particular, \citet{geshkovski2023emergence} model the evolution of token states $\{x_i(t)\}_{i=1}^N$ as a continuous-time flow
\begin{equation}
V \dot{x}_i(t) = \sum_{j=1}^N a_{i,j}(t)\, V x_j(t),
\end{equation}
where the attention weights $a_{i,j}(t)$ are typically defined through dot-product similarity,
\begin{equation}
a_{i,j}(t) =
\frac{\exp(\langle Qx_i(t), Kx_j(t)\rangle)}
{\sum_{\ell=1}^N \exp(\langle Qx_i(t), Kx_\ell(t)\rangle)}.
\end{equation}
Under this formulation, tokens behave as interacting particles whose trajectories are mediated by globally normalized similarity scores. While such systems may exhibit clustering behavior under specific spectral conditions on the value matrix $V$, such behavior is highly sensitive to initialization and architectural details~\citep{geshkovski2023emergence}. 

More generally, global normalization induces a strong tendency toward synchronization. In the mean-field limit $N \to \infty$, \citet{chen2025quantitative} show that dot-product attention corresponds to a Wasserstein gradient flow that contracts exponentially toward a single consensus state. Consequently, the empirical token distribution $\mu_t = \frac{1}{N}\sum_{i=1}^N \delta_{x_i(t)}$ tends toward a dominant mode, reflecting global synchronization and potential representational collapse~\cite{xiao2023efficient}. From this dynamical perspective, attention sink behavior can be interpreted as a discrete manifestation of this global synchronization and representation contraction process.


\noindent\textbf{Bounded-confidence Dynamics in Krause Attention.}
Krause Attention fundamentally modifies this dynamical system by replacing global dot-product coupling with distance-based, bounded-confidence interactions. Concretely, the attention weights take the form
\begin{equation}
\label{eq:krause_att}
a_{i,j}(t) =
\frac{\exp\!\left(-\frac{\|Qx_i(t)-Kx_j(t)\|^2}{2\sigma^2}\right)}
{\sum_{\ell \in \xi_i^k}
\exp\!\left(-\frac{\|Qx_i(t)-Kx_\ell(t)\|^2}{2\sigma^2}\right)},
\qquad j \in \xi_i^k,
\end{equation}
where $\xi_i^k$ denotes a selected sparse, local neighborhood associated with token $i$. 

This construction induces a \emph{bounded-confidence interaction structure}: tokens influence each other only when they are sufficiently close in representation space and belong to a shared local support. As a result, the time evolution of token states no longer follows a globally coupled flow, but instead decomposes into locally interacting subsystems.

\noindent\textbf{Asymptotic Multi-cluster Behavior.} This structural modification influences the long-term behavior of the dynamics.  
By attenuating long-range interactions, Krause Attention reduces the global synchronization effects characteristic of standard self-attention and instead \emph{encourages} the formation of multiple locally coherent groups. Under suitable choices of the scale parameter~$\sigma$ and sparsity support~$\xi_i^k$, the empirical distribution
$\mu_t$ can \emph{tend toward} a multi-atomic structure $\mu_t \;\rightharpoonup\; \sum_{k=1}^m \pi_k \delta_{\mathcal{L}_k}$
where $\mathcal{L}_k$ denote stable cluster representatives and $\pi_k$ their relative masses. This behavior closely parallels the classical Krause consensus model, where agents self-organize into multiple invariant opinion groups instead of converging to a global average. Crucially, in Krause Attention this multi-cluster tendency is induced by the \textbf{architectural constraints of the attention mechanism itself}, rather than any implicit biases arising from depth, initialization, or optimization-specific effects.


\noindent\textbf{Implications.} By encoding locality and selective interactions into the design, Krause Attention turns clustering from \emph{a fragile, emergent phenomenon} into a more \emph{stable architectural inductive bias}. This helps preserve token diversity and improve robustness against representation collapse. 

Appendix~\ref{app:cluster_proofs} provides supporting theoretical analyses that illuminate this multi-cluster behavior from two complementary viewpoints: clustering-emergence dynamics~\citep{geshkovski2023emergence} and mean-field Wasserstein gradient flows~\citep{chen2025quantitative}. These analyses characterize the behavior of Krause Attention under appropriate bounded-confidence and separation assumptions, offering mechanistic insight into why multiple clusters can arise.

\section{Experiments}

\subsection{Experimental Setup}


\textbf{Image Recognition.} We adopt the Vision Transformer~\citep{dosovitskiy2021image,touvron2021training,liu2021swin} with Krause Attention replacing the self-attention, and assess performance on CIFAR-10/100~\citep{krizhevsky2009learning}, Fashion-MNIST~\citep{xiao2017fashion}, and ImageNet-1K~\citep{deng2009imagenet}. 

\textbf{Image Generation.} We integrate Krause Attention into autoregressive Transformers ~\citep{parmar2018image} and evaluate on MNIST~\citep{lecun2002gradient} and CIFAR-10.

\textbf{Language Modeling.} We incorporate Krause Attention into LLMs as an \emph{auxiliary shortcut} rather than a replacement for self-attention. \footnote{Since long-range connections are known to be highly valuable in language modeling, this setting is intended to examine the \emph{compatibility and complementary benefits} of Krause Attention in LLMs rather than to position it as a full architectural alternative.} To demonstrate the effectiveness of this approach, we conduct experiments in two settings: (i) LoRA ~\citep{hu2022lora} finetuning of Qwen \citep{team2024qwen2} and Llama \citep{grattafiori2024llama}, and (ii) training language models from scratch at the 100M and 200M parameter scales. We evaluate these models across a comprehensive suite of established language understanding benchmarks, including BoolQ~\citep{clark2019boolq}, CB~\citep{de2019commitmentbank}, PIQA~\citep{bisk2020piqa}, MNLI~\citep{williams2018broad}, ANLI R1-R3~\citep{nie2020adversarial}, MMLU-Pro~\citep{wang2024mmlu}, IFEval~\citep{zhou2023instruction}, LAMBADA (OpenAI)~\citep{paperno2016lambada}, CBT~\citep{hill2015goldilocks}, Blimp~\citep{warstadt2020blimp}, Hellaswag~\citep{zellers2019hellaswag}, and ARC-Easy~\citep{clark2018think}. 


\textbf{Additional Details.} Across all tasks, standard Transformers serve as primary baselines. Krause Attention is mainly controlled by two task-dependent hyperparameters: the local neighborhood size $\mathcal{N}_i$ and the top-$k$ sparsity level. Additional experimental results, implementation details, and ablation studies are provided in Appendix~\ref{app:experiments}.

\subsection{Vision Transformers for Image Recognition}

\noindent\textbf{Krause Vision Transformers.} We construct Krause Vision Transformers (KViTs) by replacing self-attention in standard ViT architectures with Krause Attention, leaving all other architectural components unchanged. For small- and medium-scale datasets, we evaluate ViT-Tiny/Small/Base backbones. For ImageNet-1K, we consider the commonly used ViT-S/B families. We additionally study Krause Attention within SwinTransformers~\citep{liu2021swin}. Results for KViTs on CIFAR and ImageNet-1K are reported here.



\noindent\textbf{Results on CIFAR-10/100.} On CIFAR-10 and CIFAR-100, we configure Krause Attention with a local spatial neighborhood corresponding to a $4$-neighbor grid. The top-$k$ interaction parameter is scheduled linearly from 2 to 4 across layers. Tables~\ref{tab:vit_cifar10} and~\ref{tab:vit_cifar100} report classification accuracy, parameter counts, and FLOPs. Across all backbones and datasets, KViTs consistently improve accuracy while reducing FLOPs. Parameter counts remain effectively unchanged, confirming that the performance gains stem from the Krause Attention itself rather than the model capacity. Empirical results also demonstrate that integrating RoPE~\citep{su2024roformer} into Krause Attention yields consistent performance gains. We hypothesize that relative position encoding schemes are more compatible with distance-based attention mechanisms.


\begin{table}[h]
  \caption{Image classification results on \textbf{CIFAR-10}.}
  \center
  \resizebox{0.99\linewidth}{!}{
    \begin{tabular}{lcccr}
      \toprule
      \textbf{Models}  & \textbf{Accuracy(\%)}      & \textbf{\# Parameters}      & \textbf{FLOPs}  \\
      \toprule
       \textbf{ViT-T} & 90.75 & 5,362,762
 & 0.37G \\
     \rowcolor{gray!30} \textbf{KViT-T}    & \textbf{93.81} & 5,362,774
 & \textbf{0.25G}  \\
      \midrule
      \textbf{ViT-S}    & 93.33 & 21,342,346
 & 1.43G         \\
    \rowcolor{gray!30}  \textbf{KViT-S}  & \textbf{95.20} & 21,342,358
 & \textbf{0.97G}  \\
      \midrule
       \textbf{ViT-B}      & 92.45 & 85,152,010
 & 5.61G  \\
    \rowcolor{gray!30}  \textbf{KViT-B}    & \textbf{95.35} & 85,152,022
 & \textbf{3.77G} \\
      \midrule
    \textbf{ViT-B (RoPE)}  & 94.10 & 85,152,010
 & 5.61G  \\
    \rowcolor{gray!30}  \textbf{KViT-B (RoPE)}    & \textbf{95.68} & 85,152,022
 & \textbf{3.77G} \\
      \bottomrule
    \end{tabular}
    }
    \label{tab:vit_cifar10}
\end{table}

\begin{table}[h]
  \caption{Image classification results on \textbf{CIFAR-100}.}
  \center
  \resizebox{0.99\linewidth}{!}{
    \begin{tabular}{lcccr}
      \toprule
      \textbf{Models}  & \textbf{Accuracy(\%)}      & \textbf{\# Parameters}      & \textbf{FLOPs}  \\
      \midrule
       \textbf{ViT-T} & 66.07 & 5,380,132 & 0.37G \\
     \rowcolor{gray!30} \textbf{KViT-T}  & \textbf{74.34} & 5,380,144 & \textbf{0.25G} \\
      \midrule
      \textbf{ViT-S}    & 71.74 & 21,376,996 & 1.43G \\
    \rowcolor{gray!30}  \textbf{KViT-S}  & \textbf{77.05} & 21,377,008 & \textbf{0.97G} \\
      \midrule
       \textbf{ViT-B}      & 72.28 & 85,221,220 & 5.61G \\
    \rowcolor{gray!30}  \textbf{KViT-B}    & \textbf{78.03} & 85,221,232 & \textbf{3.77G}  \\
      \midrule
    \textbf{ViT-B (RoPE)}  & 74.84 & 85,221,220
 & 5.61G  \\
    \rowcolor{gray!30}  \textbf{KViT-B (RoPE)}    & \textbf{79.65} & 85,221,232
 & \textbf{3.77G} \\
      \bottomrule
    \end{tabular}
    }
    \label{tab:vit_cifar100}
\end{table}

\begin{figure}[h]
    \centering
    \includegraphics[width=0.99\linewidth]{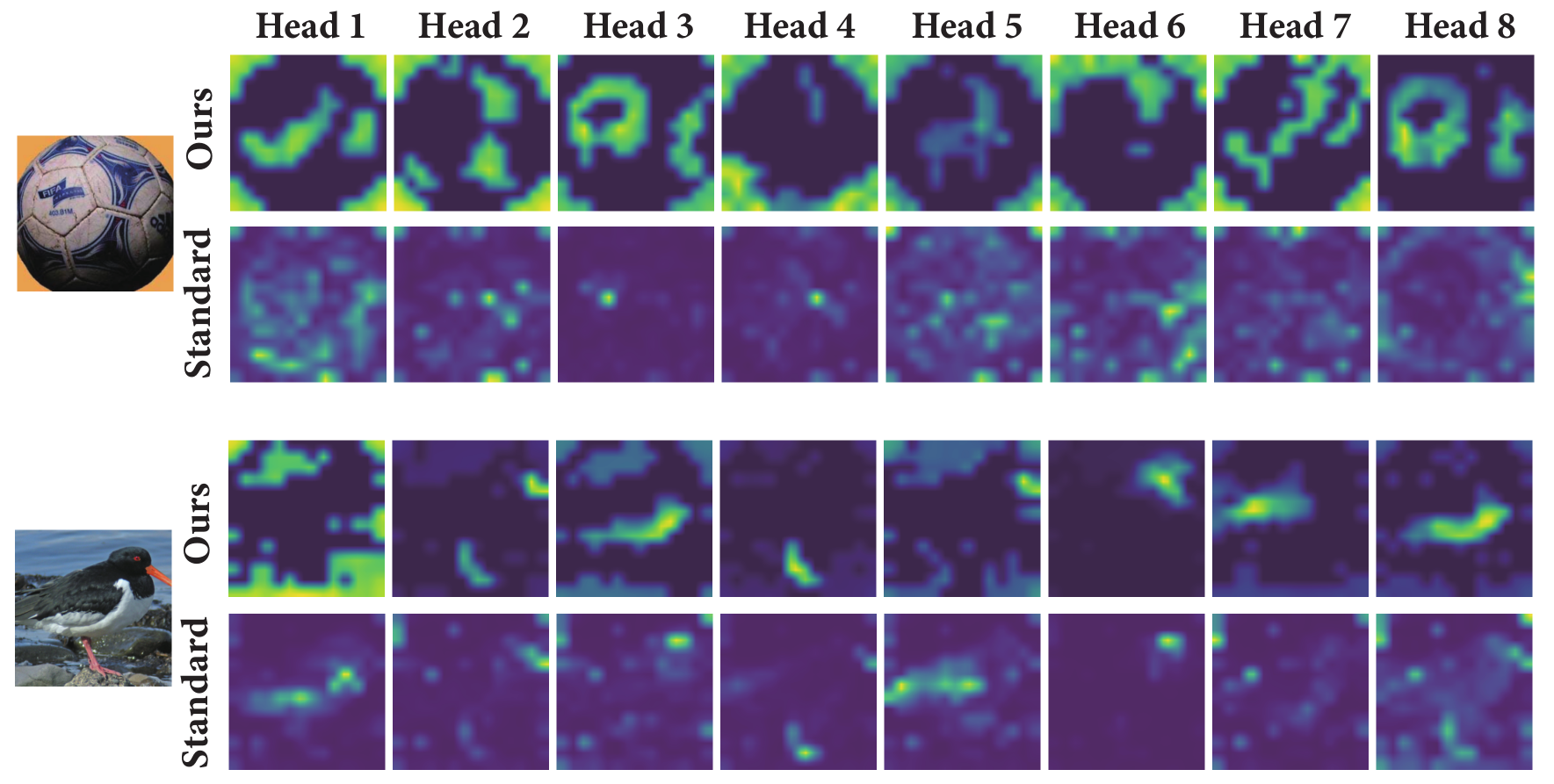}
    \caption{Krause Attention yields more diverse attention heads.}
    \label{fig:imagenet_main}
\end{figure}

\noindent\textbf{Results on ImageNet-1K.} We further evaluate KViTs on ImageNet-1K to assess performance at scale. The local interaction neighborhood is expanded to $25$ tokens, and the top-$k$ selection parameter increases linearly from $8$ to $16$ across layers to accommodate the larger spatial resolution. Table~\ref{tab:vit_imagenet1k} shows that KViTs consistently outperform ViT counterparts while reducing FLOPs. These trends hold across model sizes and patch resolutions, indicating that Krause Attention scales effectively to large-scale vision tasks. Fig.~\ref{fig:imagenet_main} further illustrates this behavior, showing that Krause Attention preserves diverse, multi-cluster attention patterns across heads, in contrast to the more homogeneous attention in ViTs.

\begin{table}[h]
  \caption{Image classification results on \textbf{ImageNet-1K}.}
  \center
  \resizebox{0.99\linewidth}{!}{
    \begin{tabular}{lcccr}
      \toprule
      \textbf{Models}  & \textbf{Accuracy(\%)}      & \textbf{\# Parameters}      & \textbf{FLOPs}  \\
      \toprule
      \textbf{ViT-S-16} & 75.54 & 22,050,664
 & 4.62G \\
      \rowcolor{gray!30}\textbf{KViT-S-16} & \textbf{76.39} & 22,050,676
 & \textbf{3.22G}  \\
      \midrule
      \textbf{ViT-S-32} & 70.66 & 22,878,952
 & 1.15G         \\
     \rowcolor{gray!30}\textbf{KViT-S-32} & \textbf{72.04} & 22,878,964
 & \textbf{0.79G}  \\
      \midrule
      \textbf{ViT-B-16} & 75.85 & 86,567,656
 & 17.61G  \\
      \rowcolor{gray!30}\textbf{KViT-B-16} & \textbf{76.75} & 86,567,668
 & \textbf{12.03G} \\
      \midrule
      \textbf{ViT-B-16 (RoPE)} & 78.40 & 86,567,656
 & 17.62G  \\
      \rowcolor{gray!30}\textbf{KViT-B-16 (RoPE)} & \textbf{78.61} & 86,567,668
 & \textbf{12.03G} \\
      \midrule
      \textbf{ViT-B-32} & 69.90 & 88,224,232
 & 4.42G \\
      \rowcolor{gray!30}\textbf{KViT-B-32} & \textbf{71.49} & 88,224,244
 & \textbf{3.00G}  \\
      \bottomrule
    \end{tabular}
    }
    \label{tab:vit_imagenet1k}
\end{table}

\noindent\textbf{Krause SwinTransformers.} To isolate the contribution of the proposed bounded-confidence interaction dynamics from architectural locality, we replace the dot-product similarity in SwinTransformers with RBF-based distance kernels in Krause Attention (KSwin-T/S), while keeping all other components unchanged. This setup provides a controlled comparison and allows us to evaluate whether distance-based interaction rules offer benefits beyond the hard-coded spatial locality already present in Swin architectures.

As shown in Table~\ref{tab:swin}, KSwin models consistently outperform their vanilla counterparts on CIFAR-10 across both model scales. Despite nearly identical parameter counts, KSwin-T/S achieve higher classification accuracy while substantially reducing FLOPs. The performance and efficiency gain arises from the kernel formulation itself, confirming that replacing similarity-based aggregation with bounded-confidence, distance-aware interactions does introduce a complementary and robust inductive bias.

\begin{table}[h]
  \caption{Image classification results of (K)Swin-T/S on \textbf{CIFAR-10}.}
  \centering
  \resizebox{0.99\linewidth}{!}{
    \begin{tabular}{lcccr}
      \toprule
      \textbf{Models}  & \textbf{Accuracy(\%)}      & \textbf{\# Parameters}      & \textbf{FLOPs}  \\
      \midrule
      \textbf{Swin-T}   & 90.73 & 27,510,484 & 0.20G  \\
      \rowcolor{gray!30}\textbf{KSwin-T}   & \textbf{91.09} & 27,510,622
 & \textbf{0.09G}  \\
      \midrule
      \textbf{Swin-S}   & 90.21 & 48,811,108 & 0.38G  \\
      \rowcolor{gray!30}\textbf{KSwin-S}   & \textbf{91.13} & 48,811,390
 & \textbf{0.18G}  \\
      \bottomrule
    \end{tabular}
    }
    \label{tab:swin}
\end{table}

\subsection{Autoregressive Models for Image Generation}

\noindent\textbf{Krause Autoregressive Models.} We further evaluate Krause Attention in the context of autoregressive image generation. Specifically, we construct Krause Autoregressive Models (KARMs) by replacing the self-attention layers in standard autoregressive Transformers with Krause Attention. We compare KARMs against two baselines: (i) standard autoregressive models with full self-attention (ARM), and (ii) autoregressive models with linear attention (LARM)~\citep{katharopoulos2020transformers}. Performance is measured using bits-per-dimension (BPD) for likelihood quality and images-per-second for inference throughput. 


\begin{table}[h]
  \caption{Image generation results of KARMs on \textbf{MNIST}. Inference speed is measured on a single NVIDIA H100 GPU.}
    \center
    \resizebox{0.99\linewidth}{!}{
    \begin{tabular}{lcccr}
      \toprule
      \textbf{Models}  & \textbf{BPD ($\downarrow$)}      & \textbf{Images/sec ($\uparrow$)}  & \textbf{Time Complexity}\\
      \midrule
      \textbf{ARM}  & 0.5685 & 83.5772  &  $\mathcal{O}(N^2d)$\\
      \textbf{LARM} & 0.5855 & 499.3672 & $\mathcal{O}(Nd^2)$\\
      \rowcolor{gray!30}\textbf{KARM}    & \textbf{0.5652} & 105.6037 & $\mathcal{O}(NWd)$\\
      \bottomrule
    \end{tabular}
    }
    \label{tab:karm_mnist}
\end{table}

\noindent\textbf{Results on MNIST.} We first conduct experiments on the MNIST dataset, where each image is represented as a sequence of length $784$. For KARMs, we employ a causal local window of size $128$ and set the top-$k$ interaction parameter to $96$ to balance locality and expressiveness. We generate $50$K samples to measure inference throughout. 

\begin{figure}[h]
    \centering
    \includegraphics[width=0.99\linewidth]{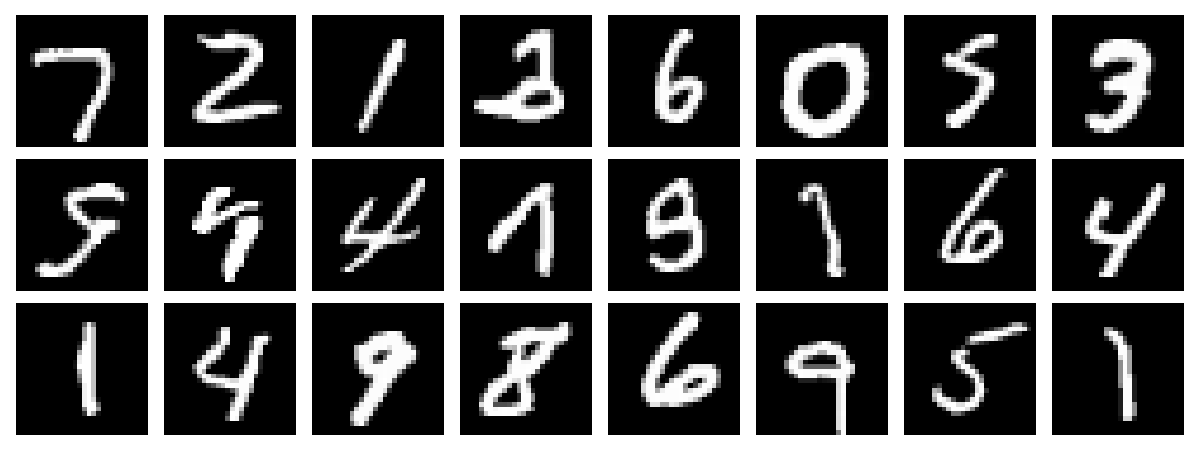}
    \caption{Unconditional samples generated by KARM on \textbf{MNIST}.}
    \label{fig:mnist_uncondition}
\end{figure}

As shown in Table~\ref{tab:karm_mnist}, KARM achieves the best likelihood (lowest BPD) among all models while also generating images substantially faster than the full-attention ARM baseline. Although KARM is slower than the linear-attention LARM, it offers markedly better likelihood, matching expectations from the intermediate time complexity $\mathcal{O}(NWd)$. Fig.~\ref{fig:mnist_uncondition} presents unconditional samples generated by KARM, illustrating the visual quality of the generated digits. 

\noindent\textbf{Results on CIFAR-10.} We next evaluate  on CIFAR-10, which corresponds to a longer sequence length of $3072$. We expand the window size to $256$ and set the top-$k$ parameter to $192$. Inference speed is evaluated on $10$K samples. 

\begin{table}[h]
  \caption{Image generation results of KARMs on \textbf{CIFAR-10}. Inference speed is measured on a single NVIDIA H100 GPU.}
    \center
    \resizebox{0.99\linewidth}{!}{
    \begin{tabular}{lcccr}
      \toprule
      \textbf{Models}  & \textbf{BPD ($\downarrow$)}      & \textbf{Images/sec ($\uparrow$)}  & \textbf{Time Complexity}\\
      \midrule
       \textbf{ARM}  & 3.0224 & 1.8933 &  $\mathcal{O}(N^2d)$\\
      \textbf{LARM} & 3.1836 & 14.4032 &  $\mathcal{O}(Nd^2)$\\
      \rowcolor{gray!30}\textbf{KARM}    & \textbf{3.0032} & 4.5240 &  $\mathcal{O}(NWd)$\\
      \bottomrule
    \end{tabular}
    }
    \label{tab:karm_cifar10}
\end{table}

Quantitative results are summarized in Table~\ref{tab:karm_cifar10}. As in the MNIST experiments, KARMs achieve the lowest BPD among all evaluated methods while significantly outperforming standard ARMs in generation speed. Although LARMs remain the fastest, they exhibit inferior likelihood performance compared to KARMs, highlighting the trade-off between efficiency and modeling accuracy.

\begin{figure}[h]
    \centering
    \includegraphics[width=0.99\linewidth]{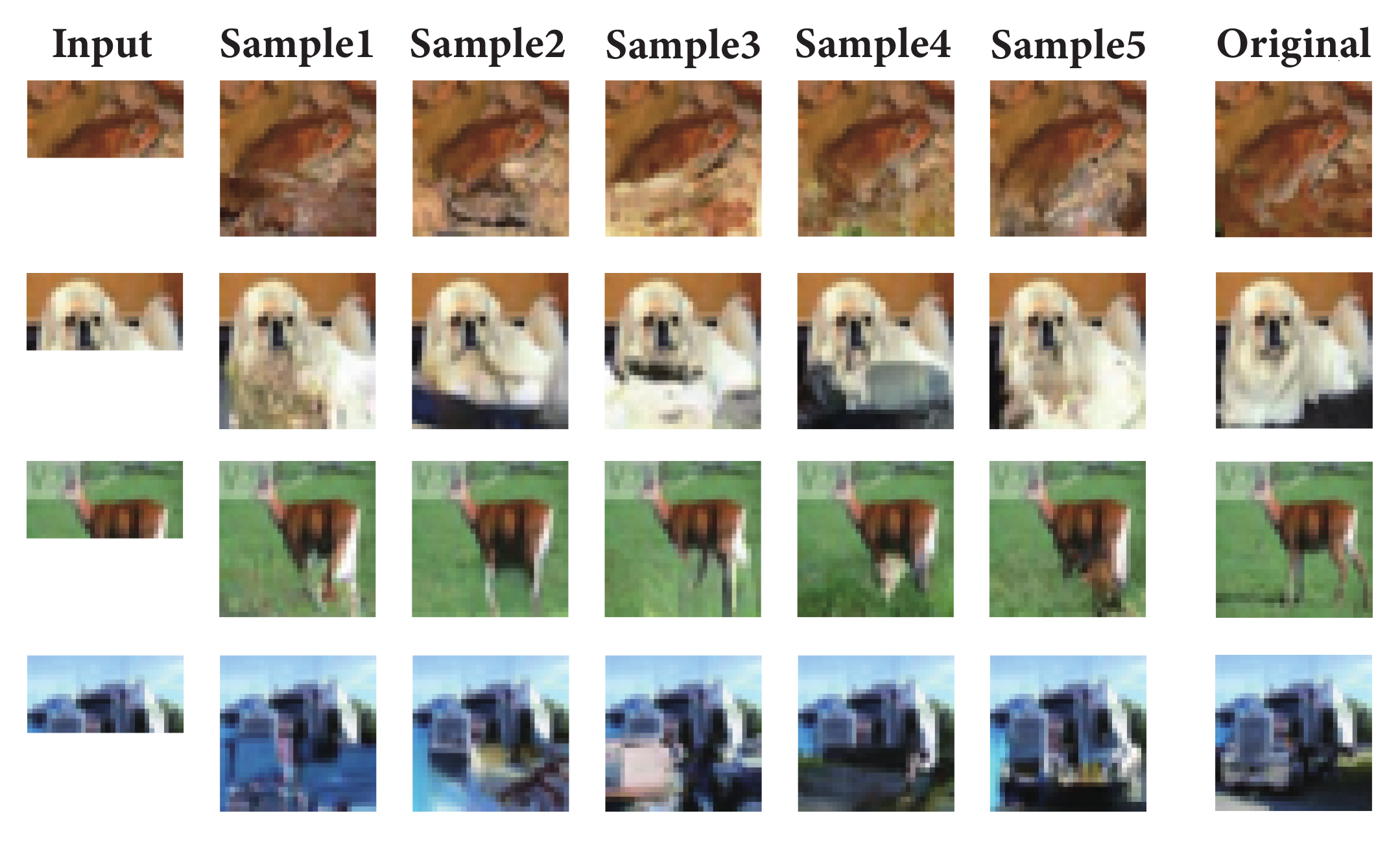}
    \caption{Samples completed by KARMs on \textbf{CIFAR-10}.}
    \label{fig:cifar10_completion}
\end{figure}

\begin{table*}[h]
    \caption{Language understanding and reasoning results of Krause-Llama3-8B. Results are reported in Acc. (\%) and Acc. / Macro-F1.}
    \center
    \small
    \setlength{\tabcolsep}{0.5pt}
    \resizebox{0.99\linewidth}{!}{
    \begin{tabular}{l|c|c|c|c|c|c|c|c|c}
      \toprule
      \footnotesize \textbf{Models}  & \textbf{BoolQ }   & \textbf{CB} & \textbf{PIQA} & \textbf{MNLI} & \textbf{ANLI-R1} & \textbf{ANLI-R2} & \textbf{ANLI-R3} & \textbf{MMLU-Pro} & \textbf{IFEval} \\
      \midrule
      \textbf{Llama3-8B} & 76.13 & 41.07/19.41 & 51.52 & 35.45/18.11 & 33.40/16.69 & 33.40/16.69 & 33.50/17.04 & 37.50 & 22.18 \\
      \textbf{Llama3-8B (finetuned w/ LoRA)} & 80.41 & 60.71/47.81 & 75.16 & 59.53/55.29 & 38.70/30.62 & 39.90/33.37 & 44.92/39.57 & \textbf{41.67} & 32.72 \\
      \midrule
      \rowcolor{gray!30}\textbf{Krause-Llama3-8B} & \textbf{80.59} & \textbf{64.29/48.04} & \textbf{77.77} & \textbf{63.27}/\textbf{53.72} & \textbf{40.30}/\textbf{33.01} & \textbf{40.50}/\textbf{34.27} & \textbf{45.67}/\textbf{39.84} & \textbf{41.67} & \textbf{34.01} \\
      \bottomrule
    \end{tabular}
    }
    \label{tab:krause_llama}
\end{table*}

We also evaluate image completion on CIFAR-10 by masking half of each input image and autoregressively generating the missing pixels. Fig.~\ref{fig:cifar10_completion} shows five completion samples generated for masked input. These results indicate that our KARMs produce semantically consistent and visually plausible completions despite the increased dataset complexity.

\subsection{LLMs for Language Understanding and Reasoning}
\label{sec:LMs}

\begin{figure}[h]
    \centering
    \includegraphics[width=0.85\linewidth]{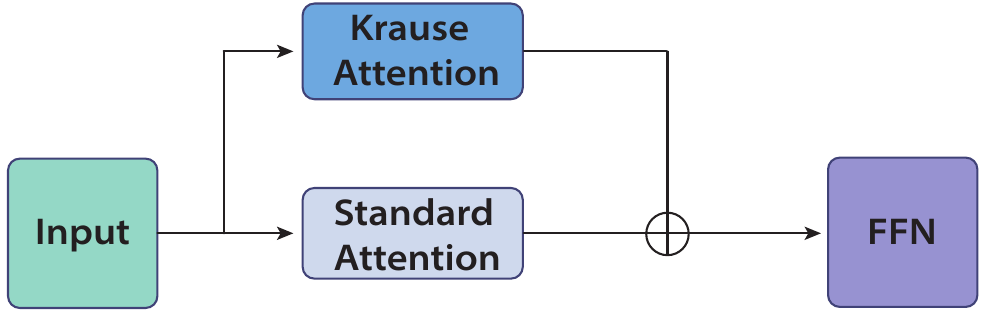}
    \caption{Our Krause models add a Krause Attention shortcut at each layer of the original LLM architecture.} 
    \label{fig:attention_mix_mechanism}
\end{figure}

\noindent\textbf{Krause-LLMs.} To explore whether Krause Attention can also improve language modeling, we integrate it into LLMs by adding a Krause-attention shortcut at every layer (Fig.~\ref{fig:attention_mix_mechanism}).  
Both the Krause pathway and the original attention modules are trained via LoRA adaptation.  
We compare this hybrid model against the base LLMs and their LoRA-finetuned variants.  
All models are trained on a 50K-sample subset of Flan-v2~\citep{longpre2023flan} and evaluated under a zero-shot setting on a suite of language benchmarks.

\noindent\textbf{Results on Krause-Llama.} For Llama3-8B, the Krause Attention is configured with a causal window size of $32$ and a top-$k$ sparsity value of $16$. Table~\ref{tab:krause_llama} presents the performance of Krause-Llama3-8B. Compared to both the base Llama3-8B and its LoRA finetuned variant, Krause-Llama3-8B achieves consistent gains on most datasets, including BoolQ, CB, PIQA, MNLI, ANLI (R1–R3), and IFEval.  
Performance on MMLU-Pro also matches the LoRA baseline.  
These results suggest that incorporating a Krause-attention shortcut can provide complementary inductive bias for language tasks without degrading overall model quality.




\noindent\textbf{Alleviating Attention Sinks.} LLMs often suffer from the \emph{attention sink} effect~\citep{xiao2023efficient}, where the \texttt{softmax} normalization allocates disproportionately high attention scores on early tokens, regardless of their semantic relevance. This behavior introduces positional bias, reduces model expressivity, and weakens representation robustness.

\begin{figure}[h]
    \centering
    \includegraphics[width=0.85\linewidth]{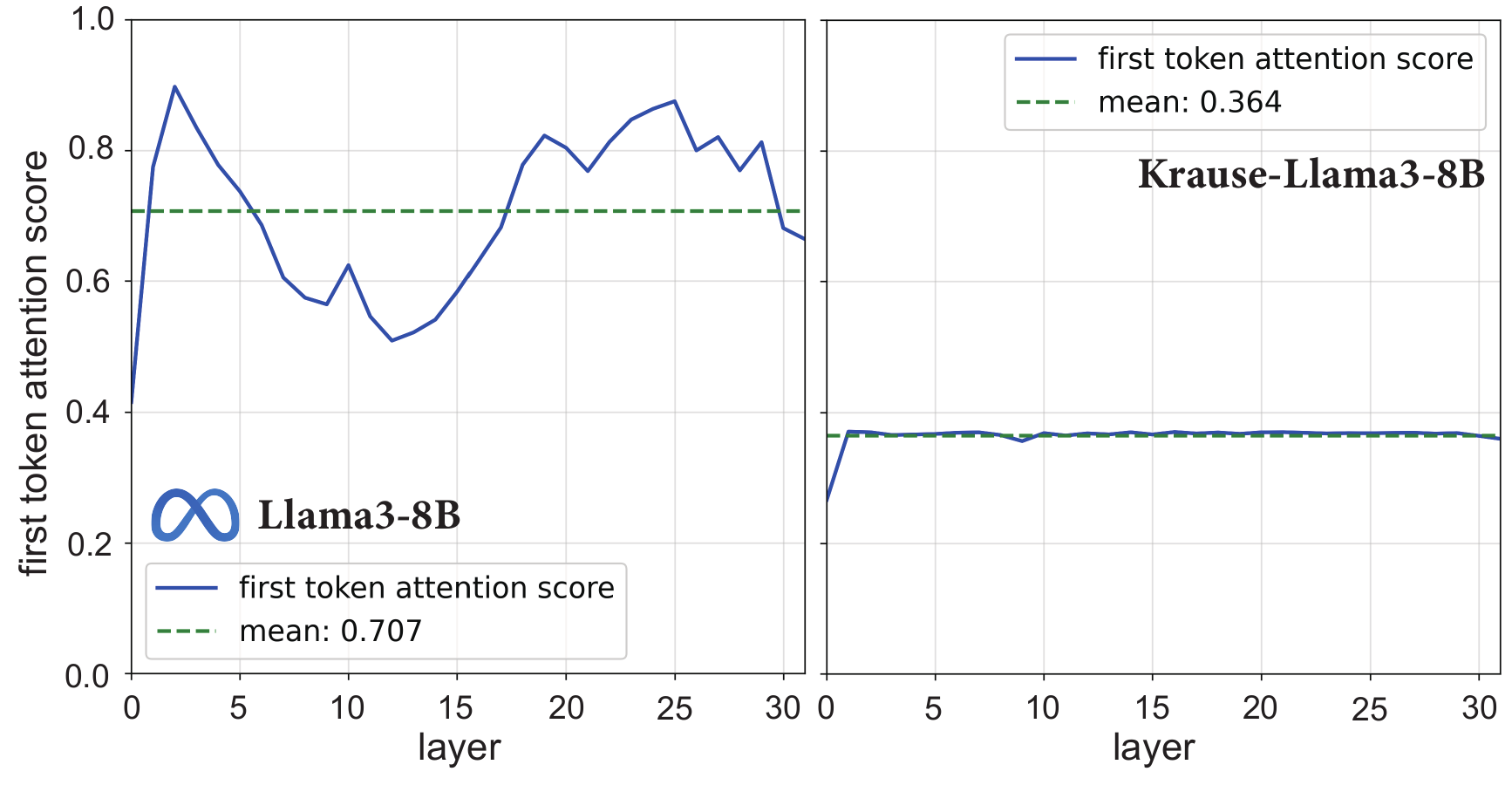}
    \caption{Layer dynamics of first-token attentions on Llama3-8B.}
    \label{fig:attention_sink_curves_llama}
\end{figure}

Krause Attention provides a complementary, bounded-confidence mechanism for mitigating this issue.  
By restricting attention to the local neighborhood, distant tokens can no longer allocate weight to the initial positions once they fall outside the receptive field.  
As shown in Fig.~\ref{fig:attention_sink_curves_llama}, the base Llama model exhibits large oscillations and persistent peaks across layers, whereas Krause-LLMs produce remarkably more stable attention curves.  
This stabilization indicates that Krause Attention reduces reliance on fixed positional anchors and supports more robust representation learning.

\noindent\textbf{Language Models Trained From Scratch.} To assess the scalability and generalization of Krause Attention, we train the 100M and 200M parameter language models from scratch on $10B$- and $20B$-token subsets of FineWeb-Edu~\citep{penedo2024fineweb}. Krause Attention is compared against several parameter-matched baselines, including sliding-window attention, standard attention, top-$k$ attention~\citep{gupta2021memory}, Longformer~\citep{beltagy2020longformer}, and the Routing Transformer~\citep{roy2021efficient}. All models are trained under the same protocol. We evaluate their performance on a suite of standard downstream benchmarks, including LAMBADA (OpenAI), CBT, PIQA, Blimp, Hellaswag, and ARC-Easy. Results are summarized in Table~\ref{tab:krause_scaling_results}. 

\begin{table}[h]
    \caption{Zero-shot performance comparison on standard downstream benchmarks at the 100M and 200M parameter scales.}
    \label{tab:krause_scaling_results}
    \centering
    \small 
    \setlength{\tabcolsep}{0.5pt}
    \begin{subtable}{\linewidth}
        \centering
        \caption{Benchmark results at the 100M parameter scale.}
        \resizebox{0.99\linewidth}{!}{
        \begin{tabularx}{\linewidth}{l|c|c|c|c|c|c}
            \toprule
            \textbf{Models} & \textbf{LAMBADA} & \textbf{CBT} & \textbf{PIQA} & \textbf{Blimp} & \textbf{Hellaswag} & \textbf{ARC-E} \\
            \midrule
            \rowcolor{gray!20}\textbf{Krause} & \textbf{24.51}  &\textbf{73.42} & \textbf{62.68} & 76.76 & 27.78 & 48.04 \\
            \textbf{Window} & \textbf{24.51} & 72.38 & 61.70 & 77.28 & 27.53 & 46.95 \\
            \textbf{Standard} & 23.33 & 72.20 & 62.62 & 77.47 & 27.37 & 47.75 \\
            \textbf{Top-$k$} & 22.80 & 71.24 & 61.81 & 75.66 & \textbf{27.95} & 48.04 \\
            \textbf{Longformer} & 22.26 & 68.18 & 61.48 & 75.22 & 27.20 & 46.99 \\
            \textbf{Routing} & 18.96 & 68.28 & 61.81 & \textbf{78.73} & 27.63 &  \textbf{49.09} \\
            \bottomrule
        \end{tabularx}
        }
        \label{tab:100m}
    \end{subtable}
    \begin{subtable}{\linewidth}
        \centering
        \caption{Benchmark results at the 200M parameter scale.}
        \resizebox{0.99\linewidth}{!}{
        \begin{tabularx}{\linewidth}{l|c|c|c|c|c|c}
            \toprule
            \textbf{Models} & \textbf{LAMBADA} & \textbf{CBT} & \textbf{PIQA} & \textbf{Blimp} & \textbf{Hellaswag} & \textbf{ARC-E} \\
            \midrule
            \rowcolor{gray!20}\textbf{Krause} & \textbf{30.60} & \textbf{80.14} & \textbf{65.07} & 79.89 & 29.81 & 51.54 \\
            \textbf{Window} & 30.45 & 78.19 & 64.25 & 79.95 & 29.81 & 52.38 \\
            \textbf{Standard} & 29.67 & 77.91 & 64.80 & 80.22 & 29.71 & 52.13 \\
            \textbf{Top-$k$} & 30.31 & 79.35 & 64.53 & 80.44 & \textbf{30.06} & 52.00 \\
            \textbf{Longformer} & 28.29 & 76.22 & 63.60 & 79.11 & 29.43 & 52.21 \\
            \textbf{Routing} & 18.84 & 70.78 & 63.87 & \textbf{80.77} & 29.45 & \textbf{52.46} \\
            \bottomrule
        \end{tabularx}
        }
        \label{tab:200m}
    \end{subtable}
\end{table}


Across both parameter scales, Krause Attention consistently outperforms all established baselines on 3 out of 6 benchmarks while remaining highly competitive on the others. These results suggest that Krause Attention constitutes a scalable and efficient alternative to prior standard, sparse, and local attention architectures.



\section{Conclusions}

This paper introduces Krause Attention, a distance-based, bounded-confidence attention mechanism that embeds an explicit multi-cluster synchronization prior into Transformer architectures. By enforcing localized and selectively sparse interactions, Krause Attention promotes structured multi-cluster token coordination, contrasting with the global synchronization tendencies observed in standard self-attention. Theoretical analyses relate Krause Attention to recent analyses of attention dynamics and to classical bounded-confidence models, offering mechanistic insight into why such architectures support multi-cluster behavior. Empirical evaluations across vision recognition, autoregressive image generation, and language models demonstrate consistent performance gains together with efficient inference. Overall, Krause Attention provides a principled and scalable inductive bias for structured token coordination, opening new avenues for dynamics-aware Transformer design.

\section*{Acknowledgments}

This research was supported in part by gifts from Cisco and OpenAI.

\section*{Impact Statement}

This paper presents work whose primary goal is to advance the field of Machine Learning by proposing a new attention mechanism grounded in bounded-confidence dynamics. Krause Attention modifies internal model interactions but does not introduce application-specific decision rules or domain-specific interventions. As with most advances in Transformer architectures, the broader societal impacts depend on the tasks and systems into which such models are deployed. While our method may contribute to improved efficiency and stability in future models, we do not identify any unique ethical concerns or societal implications beyond those already well established for research in deep learning.


\bibliography{sample}
\bibliographystyle{icml2026}


\newpage
\appendix
\onecolumn
\newpage

\section*{Krause Synchronization Transformers\\--Appendix Contents--}
\begin{itemize}
  \item {\large \textbf{\ref{app:limitation} Additional Discussion and Limitations}}
  \item {\large \textbf{\ref{app:related} Extended Related Work}}
  \item {\large \textbf{\ref{app:cluster_proofs} Multi-Cluster Behavior of Krause Attention}}
  \begin{itemize}
      \item \textbf{\ref{app:B1} Multi-Cluster Formation under Clustering Emergence Dynamics}
      \item \textbf{\ref{app:B2} Multi-Cluster Formation under Mean-Field Dynamics}
  \end{itemize}

  \item {\large \textbf{\ref{app:experiments} Extended Experiment Results}}
  \begin{itemize}
      \item \textbf{\ref{app:C1} Implementation Details}
      \begin{itemize}
          \item \ref{app:C11} Vision Transformers
          \item \ref{app:C12} Autoregressive Transformers
          \item \ref{app:C13} LLMs Finetuned with LoRA
          \item \ref{app:C14} Language Models Trained From Scratch
          \item \ref{app:C15} Evaluation Datasets and Prompts for LLMs
      \end{itemize}
    
      \item \textbf{\ref{app:C2} Vision Transformers for Image Recognition}
      \begin{itemize}
          \item \ref{app:C22} Results on Fashion-MNIST
          \item \ref{app:C23} Qualitative Results of Attention Heatmaps
          \item \ref{app:C24} Ablation Studies 
          \item \ref{app:C26} Comparison Against Extended Baselines
      \end{itemize}
    
      \item \textbf{\ref{app:C3} Autoregressive Models for Image Generation}
      \begin{itemize}
          \item \ref{app:C31} Qualitative Examples
    
          \item \ref{app:C32} Ablation Studies
      \end{itemize}
      \item \textbf{\ref{app:C4} LLMs for Language Understanding and Reasoning}
      \begin{itemize}
          \item \ref{app:C41} Results on Krause-Qwen
    
          \item \ref{app:C42} Extended Results on Krause-Llama
    
          \item \ref{app:C43} Inference Speed Comparison

          \item \ref{app:C44} Long-Context Benchmarks on Krause-Llama

          \item \ref{app:C45} Comparison Against Sink-Token Method

          \item \ref{app:C46} Results on 500M Parameter Language Models Trained From Scratch
      \end{itemize}
  \end{itemize}
  \item {\large \textbf{\ref{sec:insight} Theoretical Insights: Attention Sinks, Competitive Learning, and Over-smoothing}}
  \begin{itemize}
      \item \textbf{\ref{app:E1} Theoretical Perspective on Attention Sink Phenomena}
      \item \textbf{\ref{app:E2} Relation to Competitive Learning and Lateral Interaction Dynamics}
      \item \textbf{\ref{app:E3} Understanding Over-smoothing via Global Synchronization}
  \end{itemize}
\end{itemize}

\newpage

\section{Additional Discussion and Limitations}
\label{app:limitation}


\noindent\textbf{Theoretical Scope.}
Krause Attention is motivated by bounded-confidence and synchronization dynamics, and our theoretical analysis focuses on idealized continuous-time and mean-field formulations. As with prior dynamical analyses of Transformer models, these results characterize asymptotic behavior and provide qualitative insight into clustering and synchronization phenomena, rather than exact predictions for finite-depth, finite-width networks used in practice.

\noindent\textbf{Hyperparameter Selection.}
Krause Attention introduces several task-dependent hyperparameters, most notably the local interaction window size, the top-$k$ sparsity level, and the scale parameter $\sigma$ of the RBF distance kernel. These parameters jointly control the trade-off between locality, selectivity, and computational efficiency. While the window size and sparsity level directly affect interaction structure, the RBF scale governs the sensitivity of distance-based coupling. In practice, we observe that the scale parameter is not a critical or fragile choice: across all experiments, it reliably converges to a suitable range during training and exhibits stable behavior over a broad initialization interval. Nevertheless, we do not claim the existence of a universally optimal configuration, and systematic strategies for automatic or adaptive hyperparameter selection remain an interesting direction for future work.

\noindent\textbf{Integration with Language Models.}
In our language modeling experiments, Krause Attention is evaluated as an auxiliary interaction pathway rather than as a full replacement for standard self-attention. We validate this approach on both LoRA-finetuned LLMs and language models trained from scratch at the 100M/200M scales. Investigating deeper integration without standard attention remains an important direction for future work.


\section{Extended Related Work}
\label{app:related}

\noindent\textbf{Efficient Transformers.} Despite their empirical success, standard Transformers incur \textit{quadratic computational and memory complexity} with respect to sequence length, which fundamentally limits their scalability to long-context and high-resolution settings. To address this bottleneck, a substantial body of work has investigated sub-quadratic approximations to self-attention. One line of research reduces the cost of self-attention by imposing structured sparsity on the attention pattern. Sparse attention methods \citep{child2019generating} restrict token interactions to predefined subsets, lowering the effective complexity from $\mathcal{O}(N^2d)$ to $\mathcal{O}(N\sqrt{N}d)$.
A second class of approaches focuses on approximating the attention computation itself. Performer \citep{choromanski2021rethinking} replaces the \texttt{softmax} operation with random feature–based kernel approximations, yielding linear-time $\mathcal{O}(Nd)$ attention, while Linformer \citep{wang2020linformer} exploits low-rank structure in the attention matrix to reduce complexity. Related linear attention formulations \citep{katharopoulos2020transformers} further re-express self-attention using kernelized representations.
Finally, Reformer \citep{kitaev2020reformer} adopts a hashing-based strategy, employing locality-sensitive hashing to cluster similar tokens and obtain $\mathcal{O}(N\log Nd)$ complexity. 

\noindent \textbf{Transformer Alternatives.} Beyond the routes we discussed in the main paper, another line of research focuses on architectural alternatives. State-Space Models (SSMs), particularly Structured State-Space Sequences (S4) \citep{gu2021efficiently} and Mamba \citep{gu2023mamba,dao2024transformers}, offer linear-time inference and demonstrate superior scaling in language and vision tasks \citep{zhu2024vision,yang2024plainmamba,liu2024vmamba,munir2025vcmamba}. 
Hybrid architectures \citep{lieber2024jamba,glorioso2024zamba,de2024griffin} attempt to combine the efficiency of SSMs with the expressivity of Transformers. Other designs, such as the Perceiver \citep{jaegle2021perceiver}, utilize latent bottlenecks to process high-dimensional inputs. Despite their efficiency, these models often trade the iterative refinement of dense attention for state compression. 

\noindent\textbf{Distance-Based, Local, and Sparse Attention.} Standard Transformer architectures capture sequence geometry by augmenting dot-product attention with relative position encoding schemes, realized through additive bias terms~\citep{shaw2018self,dai2019transformer}, scalar distance penalties~\citep{press2021train}, or rotary coordinate transformations~\citep{su2024roformer}. Complementary approaches replace the query-key dot-product similarity with query–key distance kernels~\citep{nguyen2022improving,nielsen2024elliptical}, yielding attention mechanisms that are geometrically interpretable and more robust to representational anisotropy. Another line of research introduces strong inductive biases by explicitly constraining attention to localized or sparse interactions. \citet{sukhbaatar2019adaptive} learn dynamic head-specific attention spans to smoothly interpolate between local and global contexts. \citet{beltagy2020longformer} employ a deterministic sliding-window pattern augmented with selected global tokens to efficiently model long documents. \citet{gupta2021memory} leverage sparse approximations to selectively compute attention, scaling to extended contexts without full matrix evaluation. Routing Transformers \citep{roy2021efficient} employ a cluster-based routing mechanism, assigning tokens to centroids via online $k$-means and restricting attention to intra-cluster pairs. This creates an intermediate routing structure that mediates token interactions, making them dependent on clustering quality and potentially limiting direct token-to-token connectivity. 


\section{Multi-Cluster Behavior of Krause Attention}
\label{app:cluster_proofs}
This appendix provides two complementary theoretical analyses supporting the multi-cluster behavior of Krause Attention. The first argument adapts the clustering emergence framework of \citet{geshkovski2023emergence} to our sparse, distance-based attention mechanism. The second analysis leverages the mean-field Wasserstein gradient-flow perspective developed by \citet{chen2025quantitative}. Both analyses characterize the dynamics \emph{under appropriate separation or
bounded-confidence assumptions} and provide mechanistic insight into why
Krause Attention admits stable multi-cluster configurations.

\subsection{Multi-Cluster Formation under Clustering Emergence Dynamics}
\label{app:B1}

We analyze the multi-cluster formation by adopting the interacting-particle formulation of Transformer dynamics introduced by \citet{geshkovski2023emergence}. Let $\{{z}_i(t)\}_{i=1}^N\subset\mathbb{R}^D$ denote rescaled token representations evolving according to
\begin{equation} \label{eq:krause_dynamics}
\dot{z}_i(t) = \sum_{z_j \in \mathcal{N}_i(t)} a_{i,j}(t) V z_j(t),
\end{equation}
where the attention weights $a_{i,j}(t)$ are defined by the Krause Attention rule in Eq.~(\ref{eq:krause_att}), incorporating distance-based similarity, local and selective interactions.

As will be discussed in Sec.~\ref{app:B2}, bounded-confidence dynamics naturally evolve toward well-separated clusters. 
Motivated by this phenomenon, we analyze the particle-level dynamics 
\emph{after} such fragmentation has occurred and consider the regime in which tokens 
form $m$ disconnected groups $\{\mathcal{C}_k\}_{k=1}^m$ lying outside each other's interaction ranges.
This conditional analysis is standard in clustering dynamics and allows us to characterize the resulting multi-cluster behavior precisely.

\noindent\textbf{Hard Interaction Cutoff.}  
Unlike standard self-attention, Krause Attention imposes an explicit bounded-confidence rule: tokens interact only with neighbors selected by the local top-$k$ operator.  
Hence, whenever two token groups $\mathcal{G}_1$ and $\mathcal{G}_2$ are separated such that
\begin{equation}
z_2 \notin \xi_1^k \quad \text{for all } z_1 \in \mathcal{G}_1,\; z_2 \in \mathcal{G}_2,
\end{equation}
their pairwise attention weights satisfy
\begin{equation}
a_{1,2}(t)=0 \quad \text{at all times}.
\end{equation}
The sparsity and indicator structure therefore enforce a strict cutoff in the interaction graph, preventing any influence between sufficiently distant token groups.

\noindent\textbf{Block-diagonalization of Attention Dynamics.} The transition from global to local dynamics is rooted in the \textit{reducibility} of the attention matrix. In standard self-attention, the matrix $A(t)$ is strictly positive ($A_{ij} > 0$), making it an \textit{irreducible} stochastic matrix. By the Perron-Frobenius theorem, $A(t)$ has a unique largest eigenvalue $\lambda = 1$ with a corresponding strictly positive eigenvector $\mathbf{\Lambda}$, which facilitates the global convergence toward a single cluster \citep{geshkovski2023emergence}.

In contrast, the Krause Attention interaction graph $G(t)$ decomposes into $m$ disjoint connected components $\{\mathcal{C}_k\}_{k=1}^m$. Formally, if $z_i \in \mathcal{C}_p$ and $z_j \in \mathcal{C}_q$ ($p \neq q$), the sparsity mechanism ensures $A_{ij}(t) = 0$. Consequently, $A(t)$ is a \textit{reducible} stochastic matrix, and its spectrum is the union of the spectra of its blocks:
\begin{equation}
\sigma(A(t)) = \bigcup_{k=1}^m \sigma(A_{\mathcal{C}_k}(t)).
\end{equation}
Since each $A_{\mathcal{C}_k}(t)$ is itself a row-stochastic matrix, the eigenvalue $\lambda = 1$ has a multiplicity of at least $m$. The corresponding eigenspace is spanned by vectors of the clusters $\{\mathbf{\Lambda}_{\mathcal{C}_1}, \dots, \mathbf{\Lambda}_{\mathcal{C}_m}\}$, where
\begin{equation}
(\mathbf{\Lambda}_{\mathcal{C}_k})_i = 
\begin{cases} 
1 & \text{if } i \in \mathcal{C}_k \\
0 & \text{otherwise}
\end{cases}.
\end{equation}
Then the global attention matrix $A(t)$ naturally takes a block-diagonal form without requiring external constraints:
\begin{equation}
A(t) = \begin{pmatrix}
A_{\mathcal{C}_1}(t) & 0 & \dots & 0 \\
0 & A_{\mathcal{C}_2}(t) & \dots & 0 \\
\vdots & \vdots & \ddots & \vdots \\
0 & 0 & \dots & A_{\mathcal{C}_m}(t)
\end{pmatrix}.
\end{equation}
This multiplicity prevents the global contraction of tokens. Each cluster $\mathcal{C}_k$ evolves within its own invariant subspace $\mathcal{S}_k = \text{span}\{ \mathbf{\Lambda}_{\mathcal{C}_k} \}$, and its dynamics are governed independently by:
\begin{equation}
\label{krause_evolution}
\dot{Z}_{\mathcal{C}_k}(t) = A_{\mathcal{C}_k}(t) Z_{\mathcal{C}_k}(t) V^\top.
\end{equation}
Mathematically, the absence of inter-cluster spectral coupling ensures that the velocity $\dot{z}_i$ for any $i \in \mathcal{C}_p$ is a zero-gradient direction with respect to any $z_j \in \mathcal{C}_q$, thereby rigorously sustaining the multi-cluster geometry.

\noindent\textbf{Independence from the Spectrum of $V$.} In standard self-attention, the connectivity of the attention matrix implies that the dominant eigenvector of $V$ induces global alignment across all tokens \citep{geshkovski2023emergence}. If $V$ has a dominant eigenvector $v_1$, it tends to pull all $z_i$ towards the line spanned by $v_1$. In contrast, Krause Attention decouples the dynamics: even if tokens from different clusters align locally with the same eigenmode of $V$, there is no coupling force between their centroids. Considering the block dynamics in Eq.~(\ref{krause_evolution}), the clustering structure is preserved because the cross-cluster entries $[A]_{i,j}$ remain strictly zero for $z_i \in \mathcal{C}_p, z_j \in \mathcal{C}_q$ ($p \neq q$). The linear transformations induced by $V$ act only within the subspaces of each cluster and cannot reintroduce cross-cluster interaction.

Thus, Krause Attention maintains cluster separation dynamically, enabling the persistence of multi-cluster configurations independent of the global spectral dominance of $V$.

\subsection{Multi-Cluster Formation under Mean-Field Dynamics}
\label{app:B2}

We now analyze Krause Attention using the mean-field framework of \citet{chen2025quantitative}. Let $\mu_t$ denote the empirical distribution of token representations on the unit sphere $\mathbb{S}^{d-1}$.

\noindent\textbf{Truncated Interaction Kernel.}
Our Krause Attention replaces the global dot-product kernel with a truncated RBF kernel
\begin{equation}
a_{x, y} = \exp\left( -\frac{\|Qx - Ky\|^2}{2\sigma^2} \right), \quad y \in \xi_x(y),
\end{equation}
where $\xi_x(y)$ is the selected local neighborhood described in the main text, which enforces bounded-confidence interactions consistent with the classical Krause model.
In particular, the sparse neighborhood operator $\xi_x(\cdot)$ induces a compact-support interaction kernel: there exists a radius $R>0$ such that
\begin{equation}
a_{x,y} = 0 \quad \text{whenever} \quad \|Qx - Ky\| > R .
\end{equation}

The induced mean-field dynamics satisfy the continuity equation:
\begin{equation}
\partial_t \mu_t + \mathrm{div} \left( \mu_t \mathcal{X}_{\mu_t} \right) = 0, \quad
\mathcal{X}_{\mu_t}(x) = \int_{\mathbb{S}^{d-1}} \mathcal{P}_x[y] a_{x,y} \, \mathrm{d}\mu_t(y),
\end{equation}
and correspond to a Wasserstein gradient flow maximizing the interaction energy
\begin{equation}
E_{K}[\mu] = \frac{1}{2\tilde{\beta}} \iint_{\mathbb{S}^{d-1} \times \mathbb{S}^{d-1}} a_{x,y} \, d\mu(x) d\mu(y),
\qquad \tilde{\beta} := \frac{1}{2\sigma^2}.
\end{equation}

\noindent\textbf{Decomposition into Independent Flows.}
Bounded-confidence dynamics are known to generically evolve toward fragmented configurations consisting of well-separated opinion clusters
\citep{hegselmann2002opinion,lorenz2007continuous,blondel2009krause,chen2025quantitative}. In particular, uniform or sufficiently broad initial measures are linearly unstable and tend to break into clusters with characteristic spacing on the order of~$2R$, where $R$ denotes the interaction radius~\citep{wang2017noisy}. Motivated by this phenomenon, we focus our analysis on the regime \emph{after} such fragmentation has taken place. In what follows, we \emph{assume} that at some time there exists a time $t_0 \ge 0$ such that the measure $\mu_{t_0}$ admits a decomposition
\begin{equation}
\mu_{t_0} = \sum_{k=1}^m \pi_k \mu_{t_0}^{(k)}, \qquad
\mathrm{supp}(\mu_{t_0}^{(k)}) \subset \mathcal{C}_k \subset \mathbb{S}^{d-1},
\end{equation}
where the sets $\{\mathcal{C}_k\}_{k=1}^m$ satisfy
\begin{equation}
\inf_{x \in \mathcal{C}_p,\, y \in \mathcal{C}_q}
\|Qx - Ky\| > R \quad \text{for all } p \neq q .
\end{equation}
Under this condition, the interaction kernel vanishes across clusters,
\begin{equation}
a_{x,y} = 0 \quad \forall x \in \mathcal{C}_p,\; y \in \mathcal{C}_q,\; p \neq q .
\end{equation}
Consequently, for $x \in \mathcal{C}_p$,
\begin{equation}
\mathcal{X}_{\mu_t}(x)
= \int_{\mathcal{C}_p} \mathcal{P}_x[y] a_{x,y} \, \mathrm{d}\mu_t^{(p)}(y),
\end{equation}
which implies that the velocity field is entirely determined by intra-cluster interactions.

Moreover, since the velocity field $\mathcal{X}_{\mu_t}$ is tangential to $\mathbb{S}^{d-1}$
and depends only on points within the same cluster,
the flow map associated with the continuity equation preserves the supports $\mathcal{C}_k$.
In particular, no trajectory starting in $\mathcal{C}_p$ can enter the interaction range of another cluster $\mathcal{C}_q$.

Therefore, each $\mathcal{C}_k$ is forward invariant under the mean-field dynamics,
and the global flow decomposes into $m$ independent gradient flows:
\begin{equation}
\partial_t \mu^{(k)}_t + \mathrm{div} \left( \mu_t^{(k)} \mathcal{X}_{\mu_t^{(k)}} \right) = 0,
\quad
\mathcal{X}_{\mu_t^{(k)}}(x)
= \int_{\mathbb{S}^{d-1}} \mathcal{P}_x[y] a_{x,y} \, \mathrm{d}\mu_t^{(k)}(y).
\end{equation}




\noindent\textbf{Exponential Convergence within Clusters.}
We now analyze the dynamics within a single cluster $\mathcal{C}_k$.
Since $\mathcal{C}_k$ is isolated, the evolution of $\mu_t^{(k)}$ is governed by the restriction of the Krause Attention energy $E_K$ to $\mathcal{C}_k$.
Along the flow, the energy satisfies
\begin{equation}
\frac{\mathrm{d}}{\mathrm{d}t} E_{K}[\mu^{(k)}]
= \int_{\mathbb{S}^{d-1}} \| \mathcal{X}_{\mu^{(k)}}(x) \|_2^2 \,\mathrm{d}\mu^{(k)}(x)
\geqslant 0 .
\end{equation}

Fix $d \geqslant 2$, $\tilde{\beta} > 0$, $\alpha \in [0, \pi/2)$ and $u \in \mathbb{S}^{d-1}$.
Let $S_\alpha^{+(k)}(u) \subseteq \mathbb{S}^{d-1}$ denote the spherical cap
\begin{equation}
S_\alpha^{+(k)}(u)
:= \{ x \in \mathcal{C}_k \mid \langle x, u \rangle \geqslant \cos \alpha \},
\quad k = 1,\dots,m .
\end{equation}
Assume that $\mu_0^{(k)}$ is supported on $S_\alpha^{+(k)}(u)$.

If $10(1 + \sqrt{\tilde{\beta}})\tan \alpha \leqslant 1$, then following \citet{chen2025quantitative},
the interaction energy satisfies a Polyak--Łojasiewicz inequality:
\begin{equation}
E_{K}[\delta_{\mathcal{L}_k}] - E_{K}[\mu^{(k)}]
\leqslant 10 e^{-\tilde{\beta}}
\int_{S_\alpha^{+(k)}(u)}
\| \mathcal{X}_{\mu^{(k)}}(x) \|_2^2 \,\mathrm{d}\mu^{(k)}(x) .
\end{equation}

As a consequence, the solution $\mu_t^{(k)}$ converges exponentially fast to a Dirac mass $\delta_{\mathcal{L}_k}$ supported at a local consensus point $\mathcal{L}_k \in S_\alpha^{+(k)}(u)$:
\begin{equation}
W_2(\mu_t^{(k)}, \delta_{\mathcal{L}_k})
\le 20 e^{-\tilde{\beta}} e^{-\frac{e^{\tilde{\beta}}}{20} t}
\left(
\int_{S_\alpha^{+(k)}(u)}
\| \mathcal{X}_{\mu^{(k)}}(x) \|_2^2 \,\mathrm{d}\mu^{(k)}(x)
\right)^{\frac{1}{2}} .
\end{equation}
Equivalently,
\begin{equation}
W_2(\mu_t^{(k)}, \delta_{\mathcal{L}_k})
\le C_k e^{-\lambda_k t},
\end{equation}
with constants depending on $\sigma$ and the geometry of $\mathcal{C}_k$.

\noindent\textbf{Global Multi-cluster Convergence.}
Since each component measure $\mu_t^{(k)}$ converges to a Dirac mass $\delta_{\mathcal{L}_k}$ in Wasserstein gradient flow $W_2$,
the full system converges to a fragmented consensus:
\begin{equation}
\mu_t = \sum_{k=1}^m \pi_k \mu_t^{(k)}
\;\xrightarrow{W_2}\;
\sum_{k=1}^m \pi_k \delta_{\mathcal{L}_k}.
\end{equation}
This establishes the existence of stable multi-cluster equilibria under Krause Attention.

Both the clustering-emergence mechanism and the mean-field analysis demonstrate that Krause Attention fundamentally alters Transformer dynamics:
global synchronization is replaced by structured bounded-confidence interactions, encouraging stable multi-cluster representations.
This behavior contrasts sharply with the global consensus dynamics induced by standard self-attention and provides a principled theoretical explanation for the empirical robustness of Krause Attention.

\section{Extended Experiment Results}
\label{app:experiments}

\subsection{Implementation Details}
\label{app:C1}

\subsubsection{Vision Transformers} 
\label{app:C11}

All models are trained from scratch under the same training protocol. The detailed training recipes are exhibited in Table~\ref{tab:experiments_imp_cifar} for \textbf{CIFAR-10/100}, in Table~\ref{tab:experiments_imp_fashionmnist} for \textbf{Fashion-MNIST}, and in Table~\ref{tab:experiments_imp_imagenet} for \textbf{ImageNet-1K}. 

\begin{table}[h]
\centering
\small
\begin{minipage}[t]{0.48\textwidth}
\centering
\caption{Implementation details of Vision Transformers on \textbf{CIFAR-10/100}.}
\setlength{\tabcolsep}{3pt}
\begin{tabular}{l|cccccc}
\toprule
\multirow{2}{*}{Hyperparameters} & \multicolumn{3}{c}{Models} \\
\cmidrule(l){2-4}
& (K)ViT-T & (K)ViT-S & (K)ViT-B \\
\midrule
Batch Size        & 256 & 256 & 256 \\
Optimizer         & AdamW & AdamW & AdamW \\
LR                & $3 \cdot 10^{-4}$ & $3 \cdot 10^{-4}$ & $1 \cdot 10^{-4}$ \\
LR Decay          & cosine & cosine & cosine \\
Weight Decay      & 0.05 & 0.05 & 0.05 \\
Warmup Epochs     & 10   & 10   & 10   \\
Epochs            & 300 & 300 & 300 \\
GPUs              & $1 \times \text{H100}$ & $1 \times \text{H100}$ & $1 \times \text{H100}$ \\
Sigma $\sigma$    & 2.5 & 2.5 & 2.5 \\
Dropout           & \xmark & \xmark & \xmark \\
Drop Path         & 0.1/0.0 & 0.1/0.0 & 0.1/0.0 \\
H. flip           & \cmark & \cmark & \cmark \\
Mixup Alpha       & 0.8 & 0.8 & 0.8 \\
Cutmix Alpha      & 1.0 & 1.0 & 1.0 \\
Label Smoothing   & 0.1 & 0.1 & 0.1 \\
Mixup Prob.       & 0.8 & 0.8 & 0.8 \\
\bottomrule
\end{tabular}
\label{tab:experiments_imp_cifar}
\end{minipage}
\hfill
\begin{minipage}[t]{0.48\textwidth}
\centering
\caption{Implementation details of Vision Transformers on \textbf{Fashion-MNIST}.}
\setlength{\tabcolsep}{3pt}
\begin{tabular}{l|cccccc}
\toprule
\multirow{2}{*}{Hyperparameters} & \multicolumn{3}{c}{Models} \\
\cmidrule(l){2-4}
& (K)ViT-T & (K)ViT-S & (K)ViT-B \\
\midrule
Batch Size        & 256 & 256 & 256 \\
Optimizer         & AdamW & AdamW & AdamW \\
LR                & $3 \cdot 10^{-4}$ & $3 \cdot 10^{-4}$ & $1 \cdot 10^{-4}$ \\
LR Decay          & cosine & cosine & cosine \\
Weight Decay      & 0.05 & 0.05 & 0.05 \\
Warmup Epochs     & 10   & 10   & 10   \\
Epochs            & 300 & 300 & 300 \\
GPUs              & $1 \times \text{H100}$ & $1 \times \text{H100}$ & $1 \times \text{H100}$ \\
Sigma $\sigma$    & 2.5 & 2.5 & 2.5 \\
Dropout           & \xmark & \xmark & \xmark \\
Drop Path         & 0.1/0.0 & 0.1/0.0 & 0.1/0.0 \\
H. flip           & \cmark & \cmark & \cmark \\
Mixup Alpha       & 0.8 & 0.8 & 0.8 \\
Cutmix Alpha      & 1.0 & 1.0 & 1.0 \\
Label Smoothing   & 0.1 & 0.1 & 0.1 \\
Mixup Prob.      & 0.8 & 0.8 & 0.8 \\
\bottomrule
\end{tabular}
\label{tab:experiments_imp_fashionmnist}
\end{minipage}
\end{table}

\begin{table}[h]
\centering
\small
\centering
\caption{Implementation details of Vision Transformers on \textbf{ImageNet-1K}.}
\setlength{\tabcolsep}{6pt} 
\begin{tabular}{l|ccccc}
\toprule
\multirow{2}{*}{Hyperparameters} & \multicolumn{3}{c}{Models} \\
\cmidrule(l){2-4}
& KViT-S-16(32) & KViT-B-16(32) & ViT-S/B-16(32) \\
\midrule
Batch Size         & 1024 & 1024 & 1024 \\
Optimizer         & AdamW & AdamW & AdamW \\
LR                & $5 \cdot 10^{-4}$ & $5 \cdot 10^{-4}$ & $ 1 \cdot 10^{-3}$ \\
LR Decay          & cosine & cosine & cosine \\
Weight Decay      & 0.05 & 0.05 & 0.05 \\
Warmup Epochs     & 10  & 10 & 10 \\
Epochs            & 300 & 300 & 300 \\
GPUs              & $2 \times \text{H100}$ & $4 \times \text{H100}$ & $2  \times \text{H100}$ \\
Sigma $\sigma$    & 4.5 & 9.0 & \xmark \\
Dropout           & \xmark & \xmark & \xmark \\
Drop Path         & 0.1 & 0.1 & 0.0 \\
H. flip           & \cmark & \cmark & \cmark \\
Mixup Alpha       & 0.8 & 0.8 & 0.8  \\
Cutmix Alpha      & 1.0 & 1.0 & 1.0 \\
Label Smoothing   & 0.1 & 0.1 & 0.1 \\
Rand Augment      & 2\&9 & 2\&9 & 2\&9 \\
Random Erasing    & 0.25 & 0.25 & 0.25 \\
Mixup Prob.       & 0.8 & 0.8 & 0.8 \\
\bottomrule
\end{tabular}
\label{tab:experiments_imp_imagenet}
\end{table}

\begin{table}[h]
\centering
\caption{Implementation details of autoregressive Transformers.}
\setlength{\tabcolsep}{12pt}
\begin{tabular}{l|cc}
\toprule
\multirow{2}{*}{Hyperparameters} & \multicolumn{2}{c}{Models} \\
\cmidrule(l){2-3}
& \textbf{MNIST} & \textbf{CIFAR-10} \\
\midrule
Training Batch Size  & 64 & 8 \\
Optimizer         & AdamW & AdamW \\
LR                & $1 \cdot 10^{-3}$ & $1 \cdot 10^{-3}$ \\
LR Decay          & cosine & cosine \\
Warmup Ratio      & 0.05  & 0.05  \\
Epochs            & 30 & 30 \\
GPUs              & $4 \times \text{H100}$ & $4 \times \text{H100}$ \\
Sigma $\sigma$    & 2.5 & 2.5 \\
Dropout           & 0.1 & 0.1 \\
Hidden Dimension  & 256 & 512 \\
Heads             & 8 & 8 \\
Layers            & 12 & 16 \\
Temperature       & 1.0 & 1.0 \\
Generation Batch Size  & 2000 & 200 \\
\bottomrule 
\end{tabular}
\label{tab:experiments_imp_imagetf}

\end{table}


\subsubsection{Autoregressive Transformers} 
\label{app:C12}

All models are trained from scratch under the same training protocol for each dataset. Table~\ref{tab:experiments_imp_imagetf} presents the detailed configurations.


\subsubsection{LLMs Finetuned with LoRA} 
\label{app:C13}

The original LLMs as well as Krause-LLMs are finetuned under the same set of LoRA hyperparameters. We finetune the $Q, K, V, O$ projection matrices and $\texttt{MLP}$ modules for both attention mechanisms. For Krause-LLMs, at the skip connection of every layer, Krause attention is combined with standard attention with a learnable mixing ratio, which is initialized with $0.8:0.2$ across all models. Under this setup, LoRA finetuning increases the parameter count of the baseline LLMs by $0.52\%$, while Krause-LLMs incur a slightly higher overhead of $0.59\%$. The detailed settings are shown in Table~\ref{tab:lora_training_config}. 

\begin{table}[h]
\centering
\small
\caption{Implementation details of finetuning with LoRA, where $Q_{K}, K_{K}, V_{K}, O_{K}$ is the parameter set of Krause Attention.}
\setlength{\tabcolsep}{18pt} 
\begin{tabular}{lc}
\toprule
\textbf{Hyperparameters} & \textbf{Value} \\
\midrule
LoRA Rank ($r$)           & 16 \\
LoRA Alpha ($\alpha$)      & 32 \\
LoRA Dropout               & 0.1 \\
(Krause) Sigma $\sigma$     & 5.5 \\
Target Modules             & $Q, K, V, O$ projs, Gate, Up, Down projs \\
(Krause) Target Modules    & $Q, K, V, O$ projs, Gate, Up, Down projs, $Q_{K}, K_{K}, V_{K}, O_{K}$ projs \\
Task Type                  & Causal LM \\
\midrule
Optimizer                  & AdamW \\
Learning Rate (LR)         & $5 \cdot 10^{-5}$ \\
LR Scheduler               & cosine \\
Weight Decay               & 0.01 \\
Warmup Ratio               & 0.1 \\
Total Epochs               & 2 \\
Per-device Batch Size      & 4 \\
Gradient Accumulation      & 8 \\
Mixed Precision            & BF16 \\
Max Gradient Norm          & 1.0 \\
Gradient Checkpointing     & \cmark \\
\midrule
GPUs                       & $4 \times \text{H100}$ \\
Dataloader Workers         & 8 \\
Logging Steps              & 5 \\
Evaluation Strategy        & steps (150) \\
\bottomrule
\end{tabular}
\label{tab:lora_training_config}
\end{table}

\subsubsection{Language Models Trained From Scratch} 
\label{app:C14}

Krause language models and baselines are trained from scratch under the same training configuration at the 100M and 200M parameter scales. The implementation details are exhibited in Table~\ref{tab:tfsc_training_config}.

\begin{table}[h]
\centering
\small
\caption{Implementation details of training language models at both 100M and 200M parameter scales from scratch. To maintain a consistent total parameter budget across different architectures, standard and window attention are integrated via shortcut connection, and layer counts are specifically adjusted for the top-$k$ attention and routing transformer variants.}
\setlength{\tabcolsep}{18pt} 
\begin{tabular}{l|c|c}
\toprule
\textbf{Hyperparameters} & \textbf{100M} & \textbf{200M} \\
\midrule
(Krause / Window Branch) Window Size      & 32 & 32 \\
(Krause Branch / Top-$k$) Top-$k$         & 16 & 16 \\
(Krause Branch) Sigma $\sigma$     & 2.5 & 2.5 \\
(Longformer) Global Stride  & 64 & 64 \\
(Longformer) Window Size  & 256 & 256 \\
(Routing) K-means EMA Decay  & 0.99 & 0.99 \\
(Routing) Bucket Size       & 32 & 32 \\
(Routing) Dim               & 64 & 64 \\
(Routing) Clusters          & 32 & 32 \\
\midrule
Optimizer                  & AdamW & AdamW \\
Learning Rate (LR)         & $6 \cdot 10^{-4}$ & $6 \cdot 10^{-4}$ \\
LR Scheduler               & cosine & cosine \\
Weight Decay               & 0.1 & 0.1 \\
Warmup Ratio               & 0.05 & 0.05 \\
Dropout                    & 0.1 & 0.1 \\
Tokens Per Batch           & 999,424 & 999,424 \\
Total Train Tokens         & 10B & 20B \\
Mixed Precision            & BF16 & BF16 \\
Max Sequence Length        & 1024 & 1024 \\
\midrule
Hidden Size                & 768 & 768 \\
FFN Hidden Size            & 2048 & 2048 \\
Heads                      & 12 & 12 \\
Layers                     & 8 (Top-$k$/Routing 12) & 20 (Top-$k$/Routing 28) \\
\midrule
GPUs                       & $4 \times \text{H200}$ & $4 \times \text{H200}$ \\
Dataloader Workers         & 8 & 8 \\
Evaluation Strategy        & steps (500) & steps (500) \\
Save Strategy              & steps (1000) & steps (1000) \\
\bottomrule
\end{tabular}
\label{tab:tfsc_training_config}
\end{table}

\subsubsection{Evaluation Datasets and Prompts for LLMs} 
\label{app:C15}

\noindent \textbf{Evaluation Datasets.} We conduct a comprehensive evaluation of Krause-LLMs and LLMs across five critical linguistic dimensions, including reading comprehension (BoolQ), logical inference (CB, ANLI (R1-R3), and MNLI), commonsense reasoning (PIQA), expert-level knowledge (MMLU-Pro), and instruction-following evaluations (IFEval). For the IFEval benchmark, we adopt Strict Acc. as the reported evaluation metric.

We also evaluate language models trained from scratch on several standard downstream language benchmarks, including reading comprehension (CBT), linguistic competence (Blimp), commonsense reasoning (Hellaswag and PIQA), and elementary-level knowledge (ARC-E). All results are reported in Test Acc.

\noindent \textbf{Prompts for LLMs.} We present our prompts used for LoRA finetuning experiments in Table~\ref{tab:prompt_example} at the end of Appendix.

\subsection{Vision Transformers for Image Recognition}
\label{app:C2}



\begin{table}[h]
  \caption{Image classification results on \textbf{Fashion-MNIST}.}
  \centering
  \resizebox{0.5\linewidth}{!}{
        \begin{tabular}{l|cc|cc}
          \toprule
          \multicolumn{1}{c|}{\textbf{Models}} & \textbf{Accuracy (\%)} & \textbf{\# Parameters} & \textbf{FLOPs} \\
          \midrule
          \textbf{ViT-T} & 93.47 & 5,353,738 & 0.28G \\
          \rowcolor{gray!30} \textbf{KViT-T} & \textbf{94.65} & 5,353,750 & \textbf{0.19G} \\
          \midrule
          \textbf{ViT-S} & 94.11 & 21,324,298 & 1.09G \\
          \rowcolor{gray!30} \textbf{KViT-S} & \textbf{94.71} & 21,324,310 & \textbf{0.73G} \\
          \midrule
          \textbf{ViT-B} & 95.74 & 85,115,914 & 4.30G \\
          \rowcolor{gray!30} \textbf{KViT-B} & \textbf{96.10} & 85,115,926 & \textbf{2.89G} \\
          \bottomrule
        \end{tabular}
      }
      \label{tab:vit_fashionmnist}
\end{table}

\subsubsection{Results on Fashion-MNIST} 
\label{app:C22}

We further assess KViTs on Fashion-MNIST, adopting the same local neighborhood and top-$k$ settings used for CIFAR-10/100. The results in Table~\ref{tab:vit_fashionmnist} show that Krause Attention yields consistent accuracy improvements across all ViT variants, with negligible changes in parameter counts and a reduction in computational cost. These results further support the robustness and generalization benefits of Krause Attention across image datasets with different visual characteristics.

\subsubsection{Qualitative Results of Attention Heatmaps}
\label{app:C23}

\begin{figure}[h]
    \centering
    \begin{minipage}{0.49\textwidth}
        \centering
        \includegraphics[width=\linewidth]{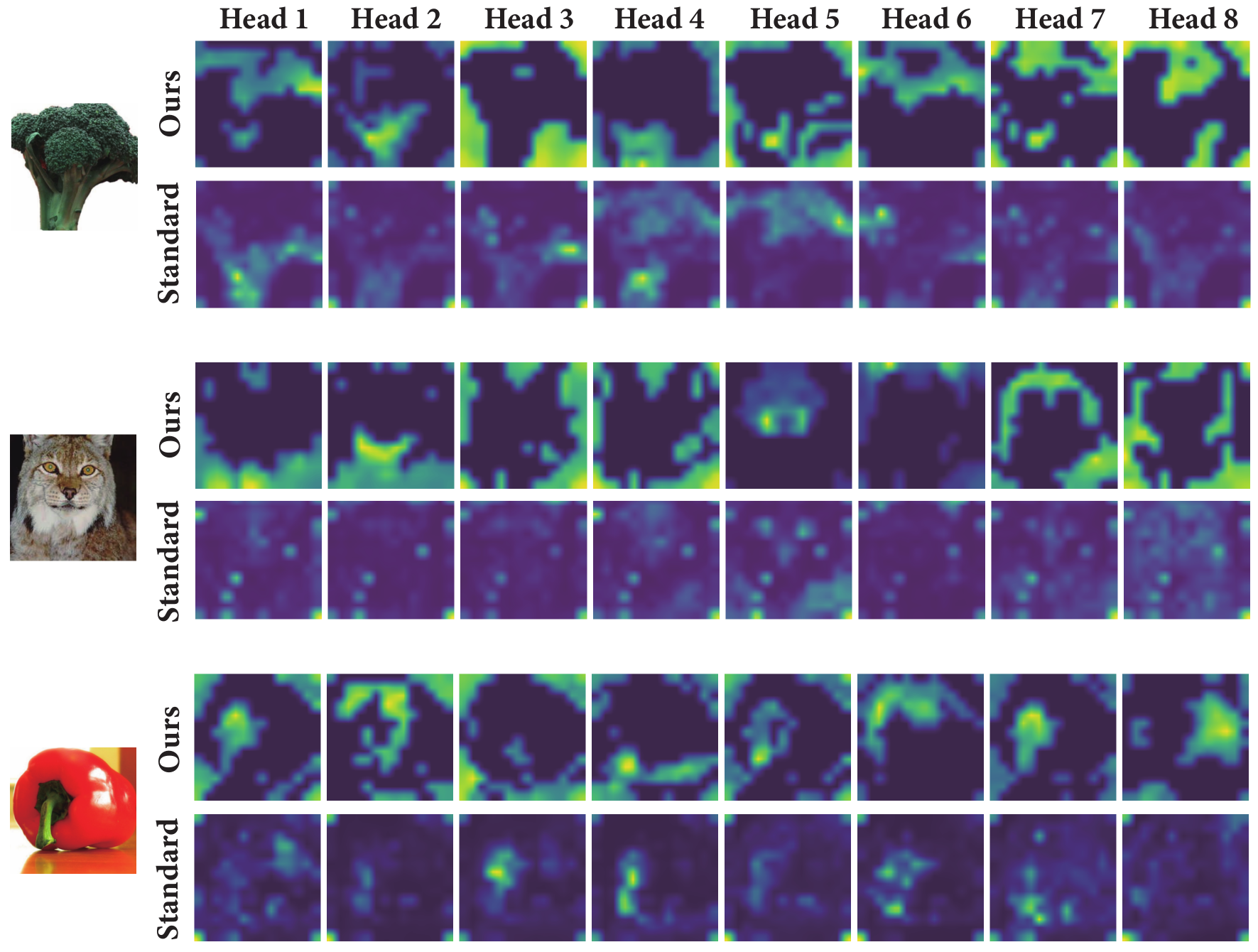}
    \end{minipage}
    \begin{minipage}{0.49\textwidth}
        \centering
        \includegraphics[width=\linewidth]{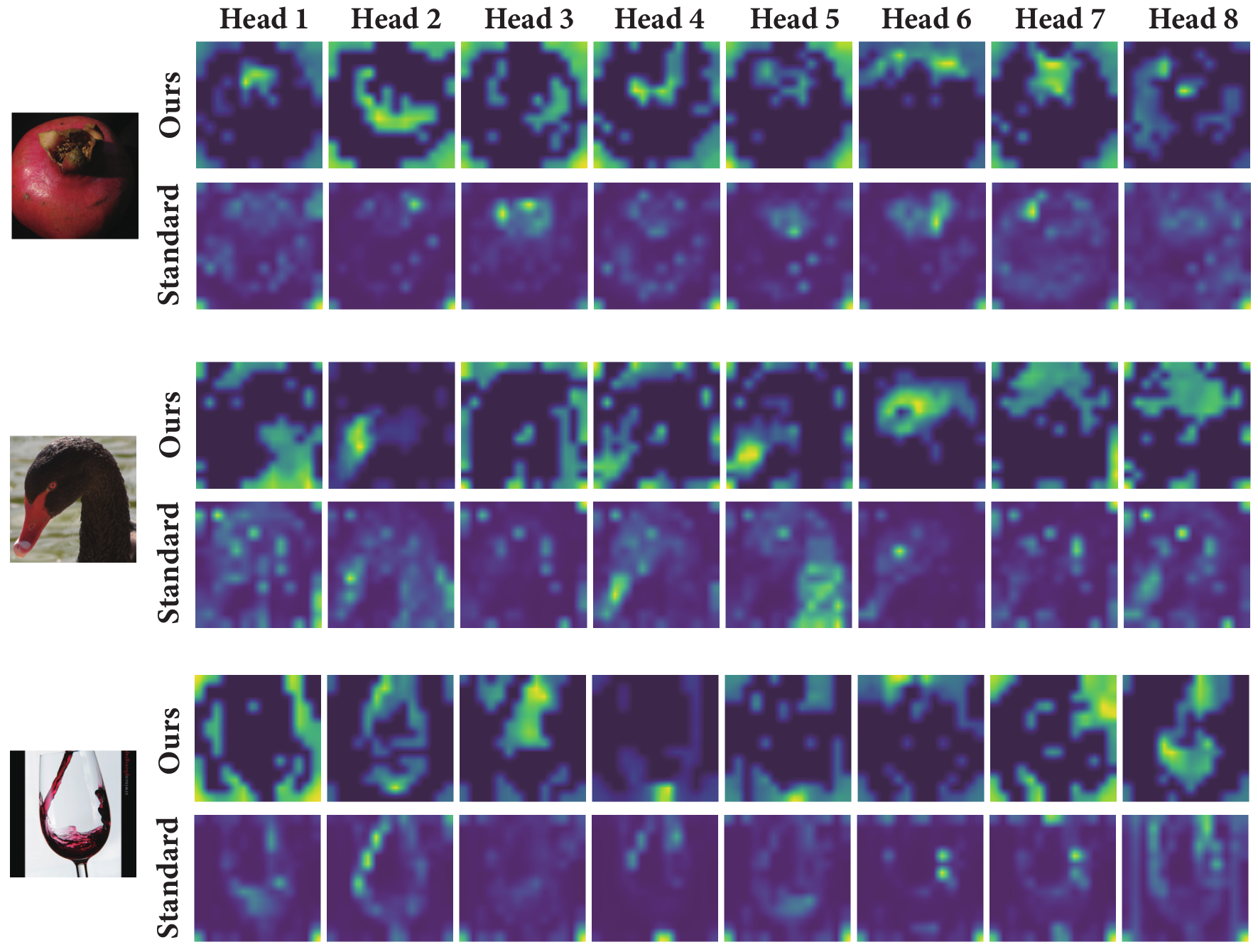}
    \end{minipage}
    \caption{Qualitative comparisons of Krause Attention and standard attention on \textbf{ImageNet-1K}.}
    \label{fig:attention_heatmaps}
\end{figure}

We visualize attention heatmaps from KViTs on representative samples in Fig.~\ref{fig:attention_heatmaps}. The results show that bounded-confidence interactions naturally give rise to distinct multi-cluster synchronization patterns across attention heads. These visualizations provide an intuitive illustration of how Krause Attention induces multiple coherent token groups rather than global mixing within Transformer layers.

\begin{figure}[h]
    \centering
    \includegraphics[width=0.99\linewidth]{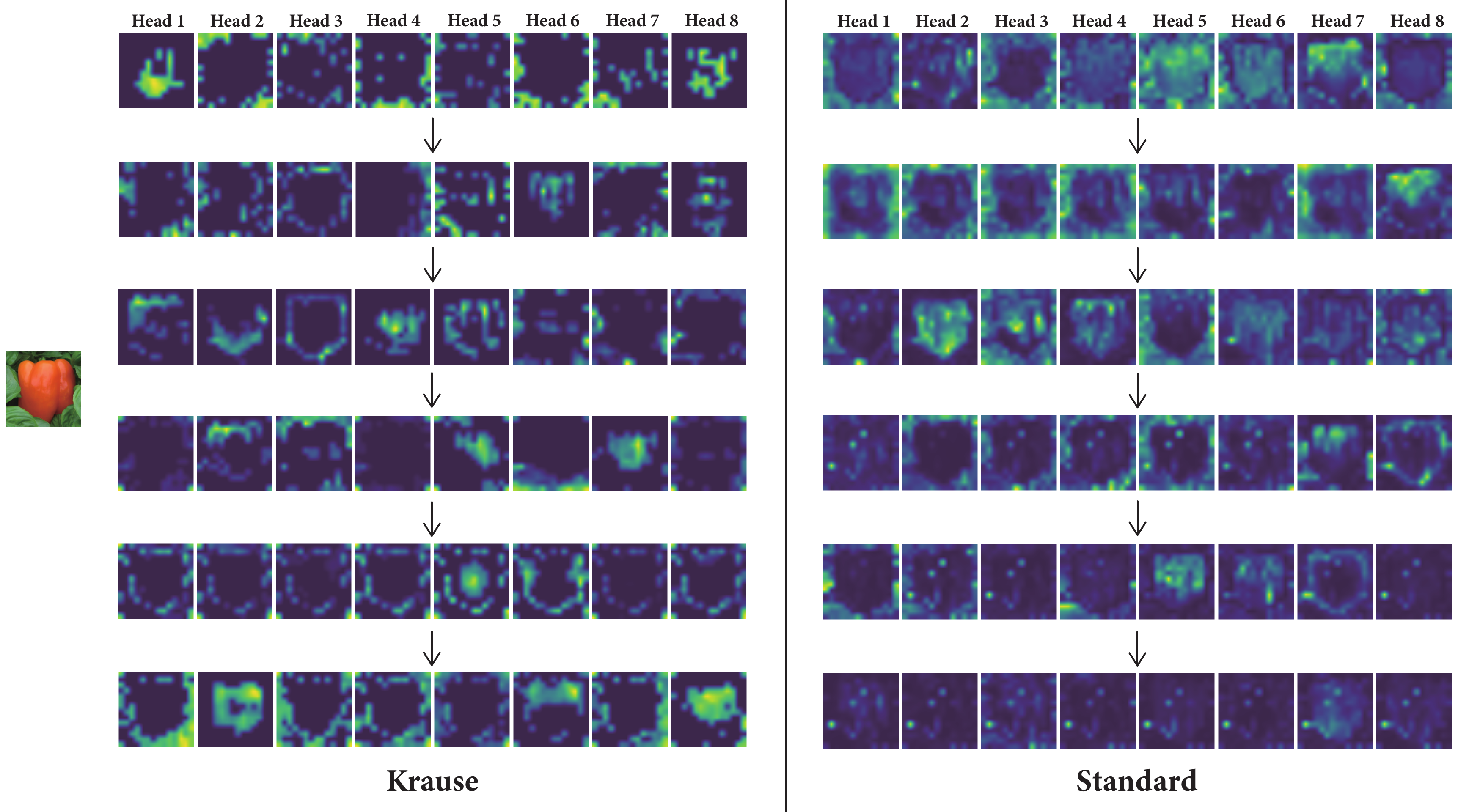}
    \caption{Evolution of attention scores across layers in KViTs/ViTs. Krause Attention (left) achieves stable multi-cluster formation, while standard attention (right) progressively converges to a single global consensus.}
    \label{fig:attention_evolution}
\end{figure}

In Fig.~\ref{fig:attention_evolution}, we further examine the evolution of attention patterns across layers. The right panel shows that standard self-attention progressively converges to nearly identical attention maps in deeper layers, a behavior often associated with representational collapse or attention sink formation. In contrast, the left panel demonstrates that Krause Attention maintains diverse, structured attention patterns throughout the network depth. This qualitative difference visually confirms the ability of Krause Attention to preserve multi-cluster structure and avoid excessive concentration of attention mass.

\subsubsection{Ablation Studies} 
\label{app:C24}

Table~\ref{tab:vit_cifar10_ablation} displays the detailed ablation studies of each component under two training settings. As can be observed, the RBF-based distance kernel used in Krause Attention already provides a substantial improvement over standard ViT-S, even in the absence of local interaction constraints or top-$k$ sparsification. This observation holds consistently across both training settings, indicating that replacing dot-product similarity with distance-based aggregation constitutes a strong inductive bias on its own.

When models are trained \emph{without} Mixup and CutMix, introducing either local interaction or top-$k$ selection leads to additional accuracy gains, and combining both mechanisms yields the best overall performance. This suggests that, in the absence of heavy data mixing, selective and localized interactions help the model focus on semantically coherent token groups and reinforce meaningful cluster formation.

In contrast, when Mixup and CutMix are applied, the benefit of top-$k$ selection becomes less pronounced. Since these augmentations partially mix image patches from different categories, the resulting token representations are less semantically localized, making selective interactions harder to exploit for improved discrimination. In this regime, enforcing top-$k$ sparsity mainly contributes to computational efficiency rather than accuracy, while local interaction remains beneficial for both performance and FLOP reduction.

Across all ablation settings, KViTs maintain essentially identical parameter counts to ViT-S. Enforcing local interaction consistently reduces FLOPs, and even the dense variant of Krause Attention remains more efficient than standard dot-product attention. Overall, these ablations disentangle the roles of the RBF kernel, locality, and selective sparsity, and highlight how their contributions interact with data augmentation strategies in shaping both performance and efficiency.

\begin{table*}[t]
  \caption{Image classification results under ablation studies on \textbf{CIFAR-10}.}
  \centering
  \resizebox{0.99\linewidth}{!}{
    \begin{tabular}{l|ccc|ccc}
      \toprule
      \multirow{2}{*}{\textbf{Models}} & \multicolumn{3}{c|}{\textbf{w/o Mixup \& Cutmix}} & \multicolumn{3}{c}{\textbf{w/ Mixup \& Cutmix}} \\
      \cmidrule(lr){2-4} \cmidrule(lr){5-7}
      & \textbf{Accuracy(\%)} & \textbf{\# Parameters} & \textbf{FLOPs} & \textbf{Accuracy(\%)} & \textbf{\# Parameters} & \textbf{FLOPs} \\
      \midrule
       \textbf{KViT-S (w/o top-$k$)}    & 91.85 & 21,342,358 & 0.97G & \textbf{95.59} & 21,342,358 & 0.97G \\
       \textbf{KViT-S (w/o local)}    & 87.55 & 21,342,358 & 1.41G & 93.96 & 21,342,358 & 1.41G \\
       \textbf{KViT-S (w/o local\&top-$k$)}  & 87.28 & 21,342,358 & 1.41G & 94.06 & 21,342,358 & 1.41G \\
       \textbf{KViT-S}   & \textbf{92.16} & 21,342,358 & 0.97G & 95.20 & 21,342,358 & 0.97G \\
      \midrule
       \textbf{ViT-S}   & 82.33 & 21,342,346 & 1.43G & 93.33 & 21,342,346 & 1.43G \\
      \bottomrule
    \end{tabular}
  }
  \label{tab:vit_cifar10_ablation}
\end{table*}

\subsubsection{Comparison Against Extended Baselines}
\label{app:C26}

Relative positional encoding schemes, ranging from early approaches like Transformer-XL~\citep{dai2019transformer} to more recent designs like RoPE~\citep{su2024roformer}, incorporate relative token distances into content-based attention computation. RoPE applies rotary transformations to query and key representations such that their dot product becomes dependent on relative positional offsets. This formulation enables the self-attention mechanism to capture long-range dependencies through the geometric structure of the feature space.

As a representative sparse attention architecture, top-$k$ attention~\citep{gupta2021memory} was introduced to alleviate the quadratic complexity of vanilla attention in sequence length. Specifically, queries are processed in chunks, and only the top-$k$ attention scores are preserved for each query. This mechanism is conceptually similar to the representation-based top-$k$ selection strategy employed in Krause Attention.

To further compare Krause Attention against both relative positional encoding schemes and sparse attention variants, we evaluate ViT-B models on CIFAR-10/100 and ImageNet-1K. We incorporate RoPE~\citep{su2024roformer} (CIFAR/ImageNet) and top-$k$ attention~\citep{gupta2021memory} (CIFAR) into ViTs to serve as representative baselines for comparison. All training protocols and hyperparameter configurations remain identical to what we used in main-text experiments. The corresponding results are reported below. 

\noindent\textbf{Results on CIFAR-10/100.} As shown in Table~\ref{tab:extended_vit_cifar10} and Table~\ref{tab:extended_vit_cifar100}, KViTs consistently outperform all baselines while reducing FLOPs. Particularly, replacing the additive positional encoding in KViTs with RoPE leads to additional accuracy improvements. We hypothesize that relative positional encoding schemes are inherently more compatible with distance-based attention mechanisms.

\begin{table}[h]
  \centering 
  \begin{minipage}[b]{0.48\linewidth}
    \centering
    \caption{Extended image classification results on \textbf{CIFAR-10}.}
    \resizebox{\linewidth}{!}{ 
      \begin{tabular}{l|cc|cc}
        \toprule
        \textbf{Models}  & \textbf{Accuracy(\%)}      & \textbf{\# Parameters}      & \textbf{FLOPs}  \\
        \toprule
         \textbf{ViT-B} & 92.45 & 85,152,010 & 5.61G  \\
         \textbf{Top-$k$ ViT-B} & 93.18 & 85,152,010 & 5.61G \\
         \textbf{ViT-B (RoPE)} & 94.10 & 85,152,010 & 5.61G \\
       \midrule
        \rowcolor{gray!30}  \textbf{KViT-B} & 95.35 & 85,152,022 & \textbf{3.77G} \\
        \rowcolor{gray!30}  \textbf{KViT-B (RoPE)} & \textbf{95.68} & 85,152,022 & \textbf{3.77G} \\
        \bottomrule
      \end{tabular}
    }
    \label{tab:extended_vit_cifar10}
  \end{minipage}
  \hfill 
  \begin{minipage}[b]{0.48\linewidth}
    \centering
    \caption{Extended image classification results on \textbf{CIFAR-100}.}
    \resizebox{\linewidth}{!}{
      \begin{tabular}{l|cc|cc}
        \toprule
        \textbf{Models}  & \textbf{Accuracy(\%)}      & \textbf{\# Parameters}      & \textbf{FLOPs}  \\
        \toprule
         \textbf{ViT-B} & 72.28 & 85,221,220 & 5.61G  \\
         \textbf{Top-$k$ ViT-B} & 76.23 & 85,221,220 & 5.61G \\
         \textbf{ViT-B (RoPE)} & 74.84 & 85,221,220 & 5.61G \\
       \midrule
        \rowcolor{gray!30}  \textbf{KViT-B} & 78.03 & 85,221,232 & \textbf{3.77G} \\
        \rowcolor{gray!30}  \textbf{KViT-B (RoPE)} & \textbf{79.65} & 85,221,232 & \textbf{3.77G} \\
        \bottomrule
      \end{tabular}
    }
    \label{tab:extended_vit_cifar100}
  \end{minipage}

\end{table}

\noindent\textbf{Results on ImageNet-1K.} We extend our evaluation to ImageNet-1K. As shown in Table~\ref{tab:extended_vit_imagenet1k}, although RoPE-based ViTs achieve higher accuracy than KViTs, incorporating RoPE into Krause Attention enables KViTs to bridge the performance gap and even surpass RoPE-based ViTs while retaining their computational efficiency advantages. These results further support our expectation that RoPE is inherently highly complementary to our distance-based attention mechanisms.

\begin{table}[h]
  \caption{Extended image classification results on \textbf{ImageNet-1K}.}
  \center
  \resizebox{0.5\linewidth}{!}{
    \begin{tabular}{l|cc|cc}
      \toprule
      \textbf{Models}  & \textbf{Accuracy(\%)}      & \textbf{\# Parameters}      & \textbf{FLOPs}  \\
      \toprule
       \textbf{ViT-B-16} & 75.85 & 86,567,656 & 17.61G  \\
       \textbf{ViT-B-16 (RoPE)} & 78.40 & 86,567,656 & 17.62G \\
      \midrule
      \rowcolor{gray!30}  \textbf{KViT-B-16} & 76.75 & 86,567,668 & \textbf{12.03G} \\
      \rowcolor{gray!30}  \textbf{KViT-B-16 (RoPE)} & \textbf{78.61} & 86,567,668 & \textbf{12.03G} \\
      \bottomrule
    \end{tabular}
    }
    \label{tab:extended_vit_imagenet1k}
\end{table}

\subsection{Autoregressive Models for Image Generation}
\label{app:C3}

\subsubsection{Qualitative Examples} 
\label{app:C31}

\begin{figure}[h]
    \centering
    \begin{minipage}{0.45\textwidth}
        \centering
        \includegraphics[width=\linewidth]{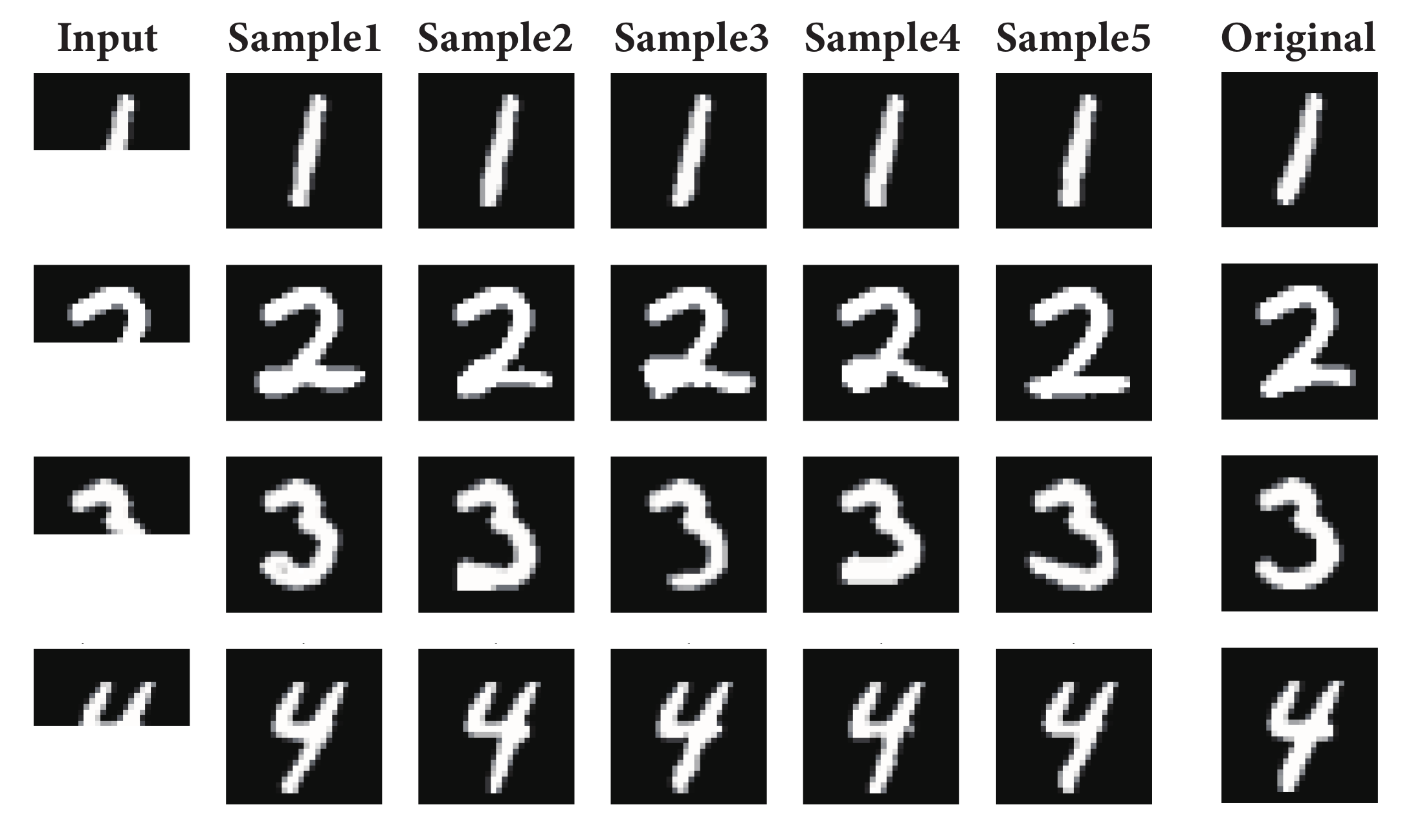}
    \end{minipage}
    \begin{minipage}{0.45\textwidth}
        \centering
        \includegraphics[width=\linewidth]{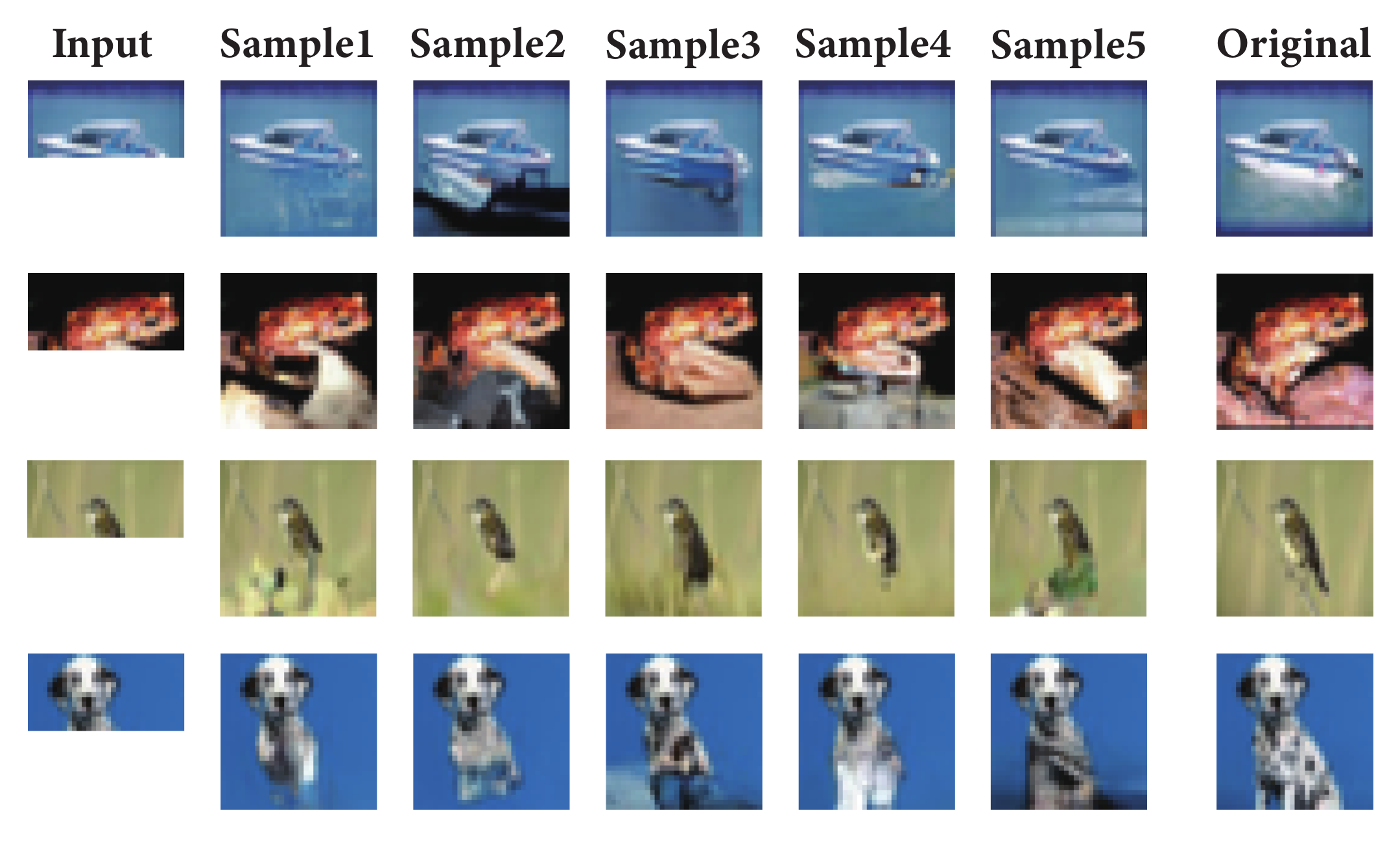}
    \end{minipage}
    \caption{Samples completed by KARMs on \textbf{MNIST} (left) and \textbf{CIFAR-10} (right).}
    \label{fig:samples_completion}
\end{figure}

We evaluate conditional generation through image completion on MNIST and CIFAR-10, where the top half of each image is masked. The model is conditioned on the visible portion and tasked with generating the remaining pixels. Representative results are shown in Fig.~\ref{fig:samples_completion}. For each input image, five completion samples are produced. These results demonstrate that KARMs generate coherent and diverse completions consistent with the underlying digit/object structure.

\subsubsection{Ablation Studies} 
\label{app:C32}

\noindent\textbf{Varying Window Size and Sparsity Level.} As shown in Table~\ref{tab:karm_mnist_varying_size}, reducing the local window size substantially increases inference throughput, with only a mild gradual degradation in generation quality. This behavior highlights a controllable trade-off between efficiency and fidelity in KARMs: smaller windows yield faster generation at the cost of slightly higher BPD, allowing practitioners to adjust the window and top-$k$ parameters according to application-specific requirements.

\begin{table}[h]
  \small
  \begin{minipage}[t]{0.46\textwidth}
    \centering
    \caption{Image generation performance of KARMs on \textbf{MNIST} with varying window sizes and sparsity levels. Inference throughput is measured on a single NVIDIA H100 GPU. Window and sparsity settings are reported as window size/top-$k$.}
    \label{tab:karm_mnist_varying_size}
    \vspace{2pt}
    \resizebox{\linewidth}{!}{
      \begin{tabular}{lccr}
        \toprule
        \textbf{Models}  & \textbf{BPD ($\downarrow$)}      & \textbf{Images/sec ($\uparrow$)} \\
        \midrule
        \textbf{KARM(128/96)}  & \textbf{0.5652} & 105.6037 \\
        \textbf{KARM(96/64)} & 0.5704 & 127.3919 \\
        \textbf{KARM(64/48)} & 0.5853 & \textbf{169.5279} \\
        \bottomrule
      \end{tabular}
    }
  \end{minipage}
  \hfill
  \begin{minipage}[t]{0.52\textwidth}
    \centering
    \caption{Ablation study of KARMs on \textbf{MNIST} image generation. Inference throughput is measured on a single NVIDIA H100 GPU.}
    \label{tab:karm_mnist_ablation}
    \vspace{10pt}
    \resizebox{\linewidth}{!}{
      \begin{tabular}{lccr}
        \toprule
        \textbf{Models}  & \textbf{BPD ($\downarrow$)}      & \textbf{Images/sec ($\uparrow$)} \\
        \midrule
        \textbf{KARM}  & 0.5652 & 105.6037 \\
        \textbf{KARM(w/o window)} & 0.5635 & 42.6981 \\
        \textbf{KARM(w/o top-$k$)} & 0.5674 & \textbf{127.7694} \\
        \textbf{KARM(w/o window\&top-$k$)} & \textbf{0.5593} & 45.9693 \\
        \bottomrule
      \end{tabular}
    }
  \end{minipage}
\end{table}

\noindent\textbf{Varying Individual Components.} Table~\ref{tab:karm_mnist_ablation} presents an ablation study that isolates the effects of the local window and top-$k$ selection in KARMs. Removing the local window improves the likelihood estimation but substantially increases per-step computation, resulting in slower overall generation. In contrast, disabling top-$k$ selection increases throughput by eliminating the selection overhead, but leads to degraded generation quality. Notably, the lowest BPD is achieved when both locality and sparsity constraints are removed, highlighting the intrinsic modeling strength of the underlying distance-based Krause Attention itself. This observation is consistent with the behavior observed in the Krause SwinTransformer experiments. 

Taken together, these results demonstrate that Krause Attention provides a modular and flexible design, where locality and sparsity can be selectively enabled to trade off performance and efficiency under different resource constraints.

\subsection{LLMs for Language Understanding and Reasoning}
\label{app:C4}

\subsubsection{Results on Krause-Qwen} 
\label{app:C41}

Qwen1.5-7B is an intermediate release in the Qwen family and shares the same architectural configuration and tokenizer as Qwen2~\citep{team2024qwen2}. We evaluate Krause-Qwen1.5-7B under a zero-shot setting across the same language understanding and reasoning benchmarks used in the main text, while systematically varying the key Krause Attention hyperparameters, namely the local window size and top-$k$ sparsity level. The results under different parameter configurations are summarized in Table~\ref{tab:krause_qwen}. Overall, Krause-Qwen1.5-7B consistently matches or improves upon the LoRA-finetuned baseline across a wide range of tasks, including BoolQ, PIQA, MNLI, ANLI, MMLU-Pro, and IFEval. Performance remains robust across different window and sparsity settings, indicating that Krause Attention is not overly sensitive to precise hyperparameter choices. Notably, on the ANLI benchmarks, LoRA finetuning of Qwen1.5-7B often leads to performance degradation. Krause-Qwen1.5-7B shows a similar trend, but with noticeably reduced degradation, indicating improved robustness of Krause attention even under potentially harmful finetuning.


\begin{table*}[h]
  \caption{Language understanding and reasoning results of Krause-Qwen1.5-7B. NLI results are reported in Acc. / Macro-F1, while other tasks are reported in Acc. (\%).  Parameter settings are reported in window size/top-$k$.}
    \center
    \small
    \setlength{\tabcolsep}{0.5pt}
    \resizebox{0.99\linewidth}{!}{
    \begin{tabular}{l|c|c|c|c|c|c|c|c|c}
      \toprule
      \footnotesize \textbf{Models}  & \textbf{BoolQ }   & \textbf{CB} & \textbf{PIQA} & \textbf{MNLI} & \textbf{ANLI-R1} & \textbf{ANLI-R2} & \textbf{ANLI-R3} & \textbf{MMLU-Pro} & \textbf{IFEval} \\
      \midrule
      \textbf{Qwen1.5-7B} & 83.80 & 76.79/53.73 & 49.57 & 57.10/46.36 & \textbf{47.30}/\textbf{38.84} & \textbf{45.40}/\textbf{39.14} & \textbf{46.67}/38.05 & 30.56 & 26.99 \\
      \textbf{Qwen1.5-7B (finetuned w/ LoRA)} & 84.08 & \textbf{87.50}/78.57 & 72.45 & 82.73/81.47 & 40.40/37.34 & 38.70/36.45 & 41.33/38.87 & 36.11 & 26.43 \\
      \midrule
      \rowcolor{gray!30}\textbf{Krause-Qwen1.5-7B (48/24)} & 84.66 & \textbf{87.50}/\textbf{78.62} & 73.42 & \textbf{83.83}/\textbf{82.82} & 40.90/37.94 & 38.20/36.18 & 41.17/38.99 & 34.72 & 31.79 \\
      \rowcolor{gray!30}\textbf{Krause-Qwen1.5-7B (32/16)} & \textbf{84.78} & 85.71/70.09 & \textbf{73.70} & 83.49/82.41 & 41.30/37.88 & 39.80/37.25 & 41.17/38.62 & \textbf{37.50} & 32.16 \\
      \rowcolor{gray!30}\textbf{Krause-Qwen1.5-7B (18/12)} & 84.63 & 85.71/70.06 & 72.66 & 83.40/82.27 & 41.20/37.64 & 39.10/36.50 & 42.33/\textbf{39.59} & \textbf{37.50} & \textbf{32.17} \\
      \bottomrule
    \end{tabular}
    }
    \label{tab:krause_qwen}
\end{table*}

\begin{figure}[h]
    \centering
    \includegraphics[width=1.0\linewidth]{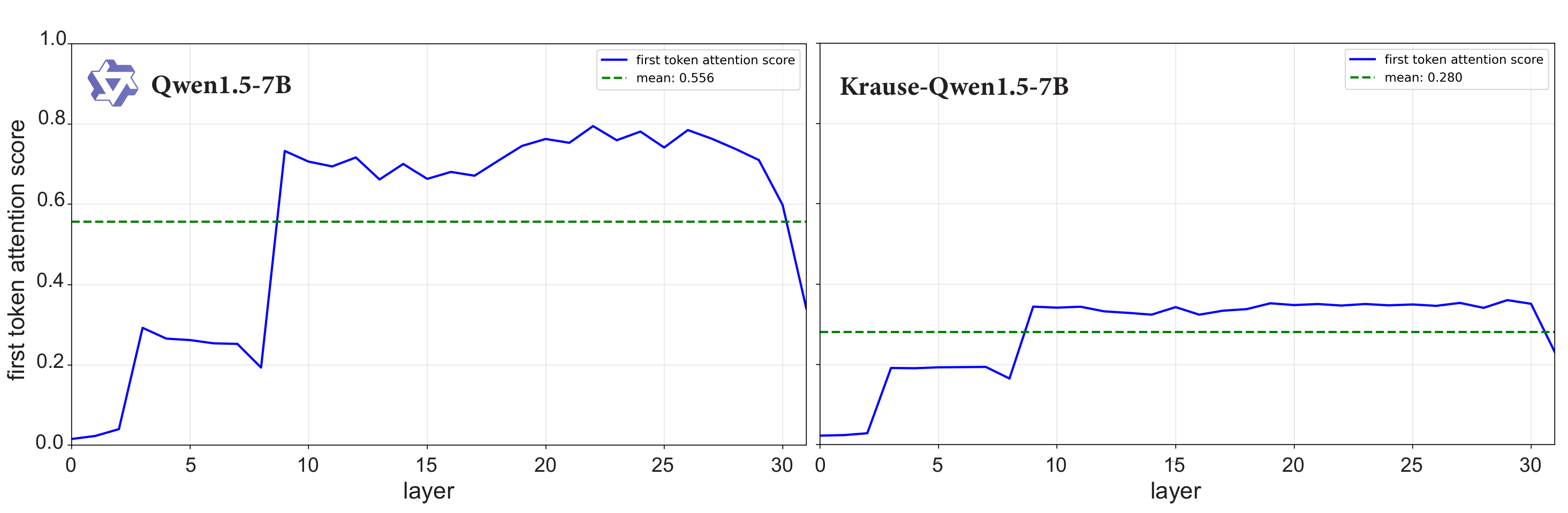}
    \caption{First token attention scores of Krause-Qwen1.5-7B (right) and Qwen1.5-7B (left).}
    \label{attention_sink_curves_qwen}
\end{figure}

Beyond task performance, Krause-Qwen1.5-7B also exhibits improved attention dynamics. Consistent with observations for Krause-Llama-3-8B, Krause-Qwen1.5-7B substantially alleviates attention sink behavior. As illustrated in Fig.~\ref{attention_sink_curves_qwen}, the vanilla Qwen1.5-7B model begins to show pronounced attention concentration on the first token starting around layer 9. In contrast, Krause-Qwen1.5-7B suppresses this collapse at comparable depths, maintaining stable and well-distributed attention scores. While recent methods such as gated attention~\citep{gate_dattention_2025} explicitly modify attention parameterization to address attention sinks, Krause Attention approaches the problem from a complementary perspective by reshaping the underlying interaction dynamics through bounded-confidence synchronization.

\subsubsection{Extended Results on Krause-Llama} 
\label{app:C42}

In addition to the results reported in the main text, we further evaluate the zero-shot performance of Krause-Llama3-8B under a range of Krause Attention configurations. Specifically, we vary the local window size and top-$k$ sparsity level while keeping all other settings fixed. The results are summarized in Table~\ref{tab:krause_llama_appendix}. Across a broad set of language understanding and reasoning benchmarks, Krause-Llama3-8B consistently outperforms the LoRA-finetuned Llama3-8B baseline, while exhibiting strong robustness across different parameter choices. Performance gains are particularly pronounced on challenging benchmarks such as ANLI and MNLI, where Krause Attention yields higher accuracy and macro-F1 scores. These results further demonstrate that the benefits of Krause Attention extend beyond a single configuration and are stable across a range of sparsity and locality hyperparameters. We also provide two examples of IFEval responses which are produced by Llama3-8B (finetuned w/ LoRA) and Krause-Llama3-8B in Table~\ref{tab:ifeval_example} at the end of Appendix.

\begin{table*}[h]
  \caption{Language understanding and reasoning results of Krause-Llama3-8B. NLI results are reported in Acc. / Macro-F1, while other tasks are reported in Acc. (\%). Parameter settings are reported in window size/top-$k$.}
    \center
    \small
    \setlength{\tabcolsep}{0.5pt}
    \resizebox{0.99\linewidth}{!}{
    \begin{tabular}{l|c|c|c|c|c|c|c|c|c}
      \toprule
      \footnotesize \textbf{Models}  & \textbf{BoolQ }   & \textbf{CB} & \textbf{PIQA} & \textbf{MNLI} & \textbf{ANLI-R1} & \textbf{ANLI-R2} & \textbf{ANLI-R3} & \textbf{MMLU-Pro} & \textbf{IFEval} \\
      \midrule
      \textbf{Llama3-8B} & 76.13 & 41.07/19.41 & 51.52 & 35.45/18.11 & 33.40/16.69 & 33.40/16.69 & 33.50/17.04 & 37.50 & 22.18 \\
      \textbf{Llama3-8B (finetuned w/ LoRA)} & 80.41 & 60.71/47.81 & 75.16 & 59.53/55.29 & 38.70/30.62 & 39.90/33.37 & 44.92/39.57 & \textbf{41.67} & 32.72 \\
      \midrule
      \rowcolor{gray!30}\textbf{Krause-Llama3-8B (48/24)} & \textbf{82.89} & \textbf{66.07}/\textbf{49.60} & 77.61 & 62.85/\textbf{56.91} & \textbf{43.20}/\textbf{39.03} & \textbf{41.30}/\textbf{37.90} & \textbf{45.83}/\textbf{42.41} & 40.28 & \textbf{36.60} \\
      \rowcolor{gray!30}\textbf{Krause-Llama3-8B (32/16)} & 80.59 & 64.29/48.04 & \textbf{77.77} & \textbf{63.27}/53.72 & 40.30/33.01 & 40.50/34.27 & 45.67/39.84 & \textbf{41.67} & 34.01 \\
      \rowcolor{gray!30}\textbf{Krause-Llama3-8B (18/12)} & 82.12 & \textbf{66.07}/47.36 & 77.50 & 62.00/54.85 & 39.50/32.25 & 40.30/34.03 & 44.83/39.64 & 38.89 & 34.57 \\
      \bottomrule
    \end{tabular}
    }
    \label{tab:krause_llama_appendix}
\end{table*}

\subsubsection{Inference Speed Comparison}
\label{app:C43}

We benchmark the inference throughput of standard LLMs and Krause-LLMs on the BoolQ dataset. As can be observed in Table~\ref{tab:inference_speed_llm}, due to the introduction of a parallel Krause Attention pathway and shortcut connections during inference, Krause-LLMs incur a modest overhead compared to their backbone models. Nevertheless, the observed throughput reduction is minimal, and Krause-LLMs maintain inference speeds that are highly comparable to both the base and LoRA-finetuned baselines across all configurations. These results indicate that the proposed attention modification introduces negligible runtime overhead while delivering consistent performance gains in language understanding tasks.

\begin{table}[h]
  \centering
  \caption{Inference speed of LLMs and Krause-LLMs on BoolQ. All results are measured on a single NVIDIA H100 GPU.}
  \label{tab:inference_speed_llm}
  \vspace{10pt} 
  \begin{minipage}{0.4\textwidth}
    \centering
    \resizebox{\linewidth}{!}{
    \begin{tabular}{lc} 
      \toprule
      \textbf{Models}  & \textbf{Tokens/sec} \\
      \midrule
      \textbf{Llama3-8B}  & 110.99 \\
      \textbf{Llama3-8B (LoRA)} & 110.88 \\
      \textbf{Krause-Llama3-8B (48/24)} & 110.10 \\
      \textbf{Krause-Llama3-8B (32/16)} & 110.26 \\
      \textbf{Krause-Llama3-8B (18/12)} & 110.64 \\
      \bottomrule
    \end{tabular}
    }
  \end{minipage}
  \hfill
  \begin{minipage}{0.4\textwidth}
    \centering
    \resizebox{\linewidth}{!}{
    \begin{tabular}{lc} 
      \toprule
      \textbf{Models}  & \textbf{Tokens/sec} \\
      \midrule
      \textbf{Qwen1.5-7B}  & 110.40 \\
      \textbf{Qwen1.5-7B (LoRA)} & 109.87 \\
      \textbf{Krause-Qwen1.5-7B (48/24)} &  109.72 \\
      \textbf{Krause-Qwen1.5-7B (32/16)} &  109.87 \\
      \textbf{Krause-Qwen1.5-7B (18/12)} &  109.84 \\
      \bottomrule
    \end{tabular}
    }
  \end{minipage}
\end{table}

\subsubsection{Long-Context Benchmarks on Krause-Llama} 
\label{app:C44}

We evaluate LoRA-finetuned Krause-Llama3-8B on LongBench~\citep{bai2024longbench}, comparing its performance against LoRA-finetuned sliding-window attention and standard attention baselines. We apply the same shortcut connection method to baselines, which is intended to ensure a comparable parameter count. As shown in Table~\ref{tab:long_context}, Krause Attention achieves the best performance on 12 out of 21 tasks and remains highly competitive on the others. This confirms that Krause Attention consistently enhances model performance across long-context benchmarks when integrated as a parallel branch alongside global attention. This improvement can be attributed to the structural advantages inherent in Krause Attention.

\begin{table*}[t]
  \caption{LongBench performance of Krause-Llama3-8B and baselines. All models are evaluated at an $8K$ context length. The configurations in the first column are denoted as window size/top-$k$.}
  \label{tab:long_context}
  \centering
  \small
  \setlength{\tabcolsep}{0.5pt}
  \begin{subtable}{0.99\linewidth}
    \centering
    \resizebox{\linewidth}{!}{
    \begin{tabular}{l|c|c|c|c|c|c|c|c|c|c}
      \toprule
      \textbf{Models} & \textbf{vcsum} & \textbf{2wikimqa} & \textbf{multi\_news} & \textbf{musique} & \textbf{multifieldqa\_en} & \textbf{hotpotqa} & \textbf{samsum} & \textbf{narrativeqa} & \textbf{lsht} & \textbf{gov\_report} \\
      \midrule
      \textbf{Llama3-8B w/ window attention} & \textbf{5.85} & \textbf{11.18} & \textbf{25.96} & 4.17 & 12.57 & 8.03 & 45.60 & 1.79 & 29.75 & \textbf{24.67} \\
      \textbf{Llama3-8B w/ standard attention} & 2.45 & 1.62 & 9.94 & 1.73 & 5.93 & 3.52 & 30.09 & 1.04 & 17.75 & 6.11 \\
      \rowcolor{gray!30}\textbf{Krause-Llama3-8B (32/16)} & 5.20 & 9.28 & 25.87 & \textbf{5.18} & \textbf{13.85} & \textbf{10.28} & \textbf{47.37} & \textbf{2.89} & \textbf{30.50} & 23.21 \\
      \bottomrule
    \end{tabular}
    }
  \end{subtable}
  \begin{subtable}{0.99\linewidth}
    \centering
    \resizebox{\linewidth}{!}{
    \begin{tabular}{l|c|c|c|c|c|c|c|c|c|c}
      \toprule
      \textbf{qmsum} & \textbf{qasper} & \textbf{dureader} & \textbf{triviaqa} & \textbf{passage\_count} & \textbf{repobench-p} & \textbf{lcc} & \textbf{multifieldqa\_zh} & \textbf{trec} & \textbf{passage\_retrieval\_zh} & \textbf{passage\_retrieval\_en} \\
      \midrule
      \textbf{13.78} & 4.64 & 7.16 & \textbf{90.34} & 1.68 & \textbf{55.76} & 53.83 & 8.43 & 71.50 & \textbf{11.37} & \textbf{7.62} \\
      2.17 & 3.43 & 5.70 & 85.67 & 1.42 & 44.55 & 41.44 & 5.89 & 71.50 & 0.50 & 1.37 \\
      \rowcolor{gray!30} 13.35 & \textbf{6.28} & \textbf{7.81} & 89.98 & \textbf{1.83} & 54.82 & \textbf{54.82} & \textbf{8.57} & \textbf{72.00} & 2.25 & 4.21 \\
      \bottomrule
    \end{tabular}
    }
  \end{subtable}
\end{table*}

We notice that while larger windows (\emph{e.g.,} 512–1K) are commonly used in sliding-window attention for long-context modeling, our design differs in that Krause Attention acts as a \emph{complementary branch} to the standard self-attention in the main pathway. The role of the Krause branch here is to capture fine-grained local interactions, rather than to replace global attention. Thus, using a smaller window can make the branch focus on modeling local structure without duplicating the role of the global pathway. We compare the Krause-Llama3-8B with different window size and top-$k$ on LongBench. Results are exhibited in Table~\ref{tab:different_window}. Interestingly, smaller window (32) consistently beats larger one (512) across most tasks, showing that in the presence of a global attention backbone, a compact local module is better than a larger-window variant.

\begin{table*}[t]
  \caption{LongBench performance of Krause-Llama3-8B with varying hyperparameters. All models are evaluated at an $8K$ context length. The configurations in the first column are denoted as window size/top-$k$.}
  \label{tab:different_window}
  \centering
  \small
  \setlength{\tabcolsep}{0.5pt}
  \begin{subtable}{0.99\linewidth}
    \centering
    \resizebox{\linewidth}{!}{
    \begin{tabular}{l|c|c|c|c|c|c|c|c|c|c}
      \toprule
      \textbf{Models} & \textbf{vcsum} & \textbf{2wikimqa} & \textbf{multi\_news} & \textbf{musique} & \textbf{multifieldqa\_en} & \textbf{hotpotqa} & \textbf{samsum} & \textbf{narrativeqa} & \textbf{lsht} & \textbf{gov\_report} \\
      \midrule
      \textbf{Krause-Llama3-8B (512/256)} & 2.66 & 8.91 & 19.82 & 4.17 & 13.33 & 8.58 & 45.88 & 1.72 & 29.00 & 16.36 \\
      \rowcolor{gray!30}\textbf{Krause-Llama3-8B (32/16)} & \textbf{5.20} & \textbf{9.28} & \textbf{25.87} & \textbf{5.18} & \textbf{13.85} & \textbf{10.28} & \textbf{47.37} & \textbf{2.89} & \textbf{30.50} & \textbf{23.21} \\
      \bottomrule
    \end{tabular}
    }
  \end{subtable}
  \begin{subtable}{0.99\linewidth}
    \centering
    \resizebox{\linewidth}{!}{
    \begin{tabular}{l|c|c|c|c|c|c|c|c|c|c}
      \toprule
      \textbf{qmsum} & \textbf{qasper} & \textbf{dureader} & \textbf{triviaqa} & \textbf{passage\_count} & \textbf{repobench-p} & \textbf{lcc} & \textbf{multifieldqa\_zh} & \textbf{trec} & \textbf{passage\_retrieval\_zh} & \textbf{passage\_retrieval\_en} \\
      \midrule
      9.67 & 3.28 & 6.08 & \textbf{90.25} & 1.73 & 52.55 & 48.26 & \textbf{11.67} & 71.00 & 0.88 & 2.43 \\
      \rowcolor{gray!30} \textbf{13.35} & \textbf{6.28} & \textbf{7.81} & 89.98 & \textbf{1.83} & \textbf{54.82} & \textbf{54.82} & 8.57 & \textbf{72.00} & \textbf{2.25} & \textbf{4.21} \\
      \bottomrule
    \end{tabular}
    }
  \end{subtable}
\end{table*}

\subsubsection{Comparison Against Sink-Token Method} 
\label{app:C45}

To demonstrate the competitiveness of Krause Attention among established attention sink remedies, we also conducted a direct comparison by finetuning Llama3-8B under identical LoRA settings using (i) a Krause Attention branch and (ii) a learnable sink token~\citep{xiao2023efficient}. Results are summarized in Table~\ref{tab:sink_token}.

\begin{table}[h]
  \caption{Language understanding and reasoning results of Krause-Llama3-8B and sink-token baseline. NLI results are reported in Acc. / Macro-F1, while other tasks are reported in Acc. (\%).  Parameter settings are reported in window size/top-$k$.}
    \center
    \small
    \setlength{\tabcolsep}{0.5pt}
    \resizebox{0.85\linewidth}{!}{
    \begin{tabular}{l|c|c|c|c|c|c|c}
      \toprule
      \footnotesize \textbf{Models} & \textbf{CB} & \textbf{PIQA} & \textbf{MNLI} & \textbf{ANLI-R1} & \textbf{ANLI-R2} & \textbf{ANLI-R3} & \textbf{MMLU-Pro} \\
      \midrule
      \textbf{Llama3-8B w/ sink token} & 58.93/41.50 & 73.32 & 56.43/49.88 & 37.20/26.34 & 37.60/28.46 & 39.92/31.52 & \textbf{47.22} \\
      \rowcolor{gray!30}\textbf{Krause-Llama3-8B (32/16)} & \textbf{64.29}/\textbf{48.04} & \textbf{77.77} & \textbf{63.27}/\textbf{53.72} & \textbf{40.30}/\textbf{33.01} & \textbf{40.50}/\textbf{34.27} & \textbf{45.67}/\textbf{39.84} & 41.67 \\
      \bottomrule
    \end{tabular}
    }
    \label{tab:sink_token}
\end{table}

Across benchmarks, the Krause-Llama3-8B consistently outperforms the sink-token approach on most tasks (CB, PIQA, MNLI, ANLI R1–R3), while remaining competitive on MMLU-Pro. This suggests that, compared to sink-token methods which reallocate attention mass within global softmax, Krause Attention provides a structural mitigation by modifying the interaction graph itself: tokens outside the local neighborhood are excluded rather than competing for probability mass. This addresses the root cause of attention sink and enforces locality across layers, leading to more stable attention patterns and improved performance.


\subsubsection{Results on 500M Parameter Language Models Trained From Scratch} 
\label{app:C46}

We train the 500M parameter language models from scratch on WikiText-103~\citep{merity2016pointer}. Following the same evaluation protocol described in the main text, we assess all models on standard downstream benchmarks. As shown in Table~\ref{tab:500m}, although the overall performance does not achieve state-of-the-art results due to the limited complexity of the training data, Krause Attention still outperforms all baselines on 4 out of 6 tasks. The results demonstrate that Krause Attention yields meaningful improvements in model quality and exhibits promising scaling potential.

\begin{table}[h]
    \centering
    \caption{Zero-shot performance comparison on standard downstream benchmarks at the 500M parameter scale.}
    \label{tab:500m}
    \small
    \begin{tabular}{l|c|c|c|c|c|c}
        \toprule
        \textbf{Models} & \textbf{LAMBADA} & \textbf{CBT} & \textbf{PIQA} & \textbf{Blimp} & \textbf{Hellaswag} & \textbf{ARC-E} \\
        \midrule
        \rowcolor{gray!20}\textbf{Krause} & \textbf{9.66} & 36.10 & \textbf{53.70} & \textbf{64.78} & \textbf{26.37} & 37.37 \\
        \textbf{Window} & 9.33 & 34.05 & 53.10 & 63.58 & 26.22 & 36.67 \\
        \textbf{Standard} & 9.26 & 38.35 & 53.48 & 64.67 & 26.32 & 36.84 \\
        \textbf{Top-$k$} & 8.25 & \textbf{40.35} & 52.61 & 64.64 & 26.29 & 37.89 \\
        \textbf{Longformer} & 7.80 & 37.25 & 53.10 & 64.13 & 26.34 & \textbf{39.12} \\
        \textbf{Routing} & 4.99 & 30.35 & 53.21 & 64.48 & 25.98 & 36.49 \\
        \bottomrule
    \end{tabular}
\end{table}

\section{Theoretical Insights: Attention Sinks, Competitive Learning, and Over-smoothing}
\label{sec:insight}

\subsection{Theoretical Perspective on Attention Sink Phenomena.}
\label{app:E1}

Recent empirical studies have documented the emergence of attention sinks in deep Transformer models~\citep{xiao2023efficient,gate_dattention_2025}, in which a small subset of tokens persistently accumulates a disproportionate fraction of attention mass across layers and heads. While this phenomenon is typically characterized at the level of attention distributions, it admits a natural interpretation from the dynamical systems viewpoint of self-attention. In particular, attention sinks can be viewed as a structural consequence of the global synchronization dynamics induced by standard self-attention.

Prior theoretical analyses~\citep{geshkovski2023emergence,chen2025quantitative} model self-attention as a globally coupled interacting particle system, where token representations evolve under repeated, globally normalized aggregation. Under dot-product attention, this interaction induces a contraction in representation space. In the mean-field limit and as depth increases, the system exhibits a strong tendency toward global synchronization, with token representations converging toward a single dominant mode. This contraction progressively reduces representational diversity and can lead to representational collapse.

From this viewpoint, \textbf{\emph{attention sinks arise as a concrete manifestation of the same collapse dynamics.}} Global softmax normalization amplifies small initial asymmetries among tokens, which may originate from positional bias, frequency effects, or stochastic initialization. Tokens that receive slightly higher attention early on become increasingly influential, shaping subsequent representations and reinforcing their dominance through repeated attention updates. This positive feedback loop translates representation synchronization into extreme concentration of attention mass, yielding a small number of persistent sink tokens. Attention sinks therefore do not constitute an independent mechanism, but emerge naturally from the global coupling and contraction dynamics of standard self-attention.

Krause Attention directly intervenes in this dynamical pathway. By replacing global dot-product similarity with bounded-confidence, distance-based interactions and enforcing locality and selective sparsity, Krause Attention prevents unrestricted global coupling. This modification breaks the positive feedback loop responsible for global synchronization and representation collapse, instead promoting stable multi-cluster dynamics. Consequently, no single token can dominate as a global attractor, providing a principled theoretical explanation for the empirically observed mitigation of attention sink behavior.


\subsection{Relation to Competitive Learning and Lateral Interaction Dynamics.}
\label{app:E2}

The clustering behavior induced by Krause Attention is closely related to classical ideas from competitive learning and lateral interaction models developed in neuroscience and early machine learning. In competitive learning frameworks, neurons or units compete to represent inputs, leading to the emergence of a small number of dominant prototypes or clusters. Canonical examples include vector quantization and self-organizing maps (SOMs), where learning dynamics drive representations toward a partitioned organization of the input space~\citep{gray1984vector,rumelhart1985feature,kohonen2002self}.

In biological neural systems, such competition is often implemented through lateral interactions within a layer. A well-studied mechanism is the Mexican-hat interaction pattern, which combines short-range excitation with longer-range inhibition \citep{amari1977dynamics,ermentrout1998neural}. At a high level, the activity $u_i$ of a neuron can be described by a lateral interaction rule:
\begin{equation}
    u_i \leftarrow \sum_j K(d_{i,j}) u_j
\end{equation}
where $d_{i,j}$ denotes the distance in feature or cortical space, and the kernel $K(\cdot)$ is positive at short range and negative at long range. This center–surround structure causes nearby neurons to synchronize and form local clusters, while distant neurons are suppressed. As a result, lateral inhibition prevents unrestricted global synchronization and instead promotes specialization, diversity, and the emergence of multiple stable activity patterns \citep{grossberg1988nonlinear}.

Krause bounded-confidence dynamics \citep{krause2000discrete,hegselmann2002opinion} can be viewed as a simplified and analytically tractable abstraction of this principle. Rather than explicitly introducing inhibitory interactions, Krause models restrict interactions to agents within a confidence radius. Agents synchronize locally with nearby neighbors, while long-range interactions are entirely removed. This implicit form of inhibition is sufficient to prevent global consensus and leads instead to robust multi-cluster equilibria, a hallmark of competitive self-organization in interacting particle systems.

Krause Attention instantiates this idea within Transformer architectures. Instead of computing attention weights from global query–key similarity, it defines interactions through query–key distances and restricts aggregation to a bounded local neighborhood. In its simplest form, Krause Attention computes
\begin{equation}
    a_{i,j} \propto \exp(-||Qx_k - Kx_j||^2),\quad j\in \xi_i,
\end{equation}
followed by normalization over a local support $\xi_i$ and optional top-$k$ selection. Nearby tokens reinforce one another through smooth, distance-aware coupling, while distant tokens are excluded from interaction altogether. 

Although Krause Attention does not introduce explicit repulsive forces as in classical Mexican-hat kernels, it implements an implicit form of lateral inhibition. Long-range interactions are suppressed through bounded-confidence constraints, while normalization and sparsity restrict competition to local neighborhoods. As a result, distant tokens neither attract nor compete with one another, which functionally mirrors the role of inhibition in preventing global synchronization in lateral-interaction models. From a dynamical viewpoint, each Krause Attention layer performs a single discrete update of a bounded-confidence interaction process, and stacking layers corresponds to iterating a competitive self-organization dynamic, analogous to repeatedly equilibrating lateral interactions in classical competitive learning systems. This interpretation aligns naturally with recent views of Transformers as interacting particle systems \citep{geshkovski2023emergence,chen2025quantitative} and clarifies why Krause Attention supports stable multi-cluster equilibria rather than collapsing toward a single global mode.

\textbf{\emph{These observations situate Krause Attention within a broader class of competitive self-organization models.}} Analogous dynamics appear in lateral-interaction systems, where local excitation and long-range inhibition yield multi-cluster equilibria. Krause Attention captures this principle in an abstract form, realizing inhibitory effects implicitly through bounded-confidence, distance-based interactions rather than explicit repulsion.

This perspective also provides insight into the mitigation of attention sink phenomena. Attention sinks can be viewed as an extreme outcome of unrestricted global competition under repeated globally normalized attention updates \citep{xiao2023efficient}. In contrast, Krause Attention confines competition to local neighborhoods, preventing any single token from emerging as a global attractor. In this sense, bounded-confidence dynamics play a stabilizing role analogous to lateral inhibition in competitive learning, preserving representational diversity while enabling structured and coherent coordination.


\subsection{Understanding Over-smoothing via Global Synchronization.} 
\label{app:E3}

Over-smoothing is a well-documented failure mode in deep architectures, manifesting as the progressive homogenization of hidden representations until token features become highly correlated and ultimately indistinguishable as model depth increases ~\citep{li2018deeper,oono2019graph,dong2021attention}. Geometrically, this corresponds to rank collapse of the representation matrix, where the feature space degenerates into a low-dimensional subspace.

From a dynamical systems perspective, over-smoothing exhibits a similar characterization with global synchronization: token trajectories converge to a single dominant attractor, collapsing the representational diversity necessary for expressive computation. This viewpoint is consistent with recent work modeling Transformers as continuous dynamical systems~\citep{nguyen2023mitigating,wang2022anti,noguchi2025wavy}. Krause Attention provides a principled countermeasure. By restricting interactions to local neighborhoods defined by feature-space proximity, it structurally suppresses the tendency towards global synchronization and promotes stable multi-cluster dynamics, preserving representational diversity and mitigating rank collapse in deep Transformers.

\begin{table}[h]
\centering
\begin{tcolorbox}[colback=gray!5.5!white,
   colframe=black!65!black, boxrule=0.3pt, rounded corners, arc=2mm]
\textbf{BoolQ Prompt Example:}\vspace{0.5em}\\
{\small \textbf{Passage}: Ethanol fuel -- All biomass goes through at least some of these steps: it needs to be grown, collected, dried, fermented, distilled, and burned. All of these steps require resources and an infrastructure. The total amount of energy input into the process compared to the energy released by burning the resulting ethanol fuel is known as the energy balance (or ``energy returned on energy invested''). Figures compiled in a 2007 report by National Geographic Magazine point to modest results for corn ethanol produced in the US: one unit of fossil-fuel energy is required to create 1.3 energy units from the resulting ethanol. The energy balance for sugarcane ethanol produced in Brazil is more favorable, with one unit of fossil-fuel energy required to create 8 from the ethanol. Energy balance estimates are not easily produced, thus numerous such reports have been generated that are contradictory. For instance, a separate survey reports that production of ethanol from sugarcane, which requires a tropical climate to grow productively, returns from 8 to 9 units of energy for each unit expended, as compared to corn, which only returns about 1.34 units of fuel energy for each unit of energy expended. A 2006 University of California Berkeley study, after analyzing six separate studies, concluded that producing ethanol from corn uses much less petroleum than producing gasoline.

\textbf{Question}: does ethanol take more energy make that produces?

\#\#\# \textbf{Response}: [Model predicts: Yes/No]}\vspace{0.5em}\\

\textbf{NLI Prompt Example:}\vspace{0.5em}\\
{\textbf{Premise}: Valence the void-brain, Valence the virtuous valet. Why couldn't the figger choose his own portion of titanic anatomy to shaft? Did he think he was helping?

\textbf{Hypothesis}: Valence was helping

A. entailment
B. contradiction
C. neutral

\#\#\# \textbf{Response}: [Model predicts token: A, B, or C]}\vspace{0.5em}\\

\textbf{PIQA Prompt Example:}\vspace{0.5em}\\
{\textbf{Question}: How do I ready a guinea pig cage for it's new occupants?

A. Provide the guinea pig with a cage full of a few inches of bedding made of ripped paper strips, you will also need to supply it with a water bottle and a food dish.

B. Provide the guinea pig with a cage full of a few inches of bedding made of ripped jeans material, you will also need to supply it with a water bottle and a food dish.

\#\#\# \textbf{Response}: [Model predicts token: A, or B]}\vspace{0.5em}\\

\textbf{MMLU-Pro Prompt Example:}\vspace{0.5em}\\
{\textbf{Question}: The symmetric group $S_n$ has $n$ elements, hence it is not true that $S_{10}$ has 10 elements.
Find the characteristic of the ring 2Z.

\textbf{Choices}:

A. 0
B. 30
C. 3
D. 10
E. 12
F. 50
G. 2
H. 100
I. 20
J. 5

\#\#\# \textbf{Response}: [Model predicts token: A-J]}\vspace{0.5em}\\

\textbf{IFEval Prompt Example:}\vspace{0.5em}\\
{Given the sentence "Two young boys with toy guns and horns." can you ask a question? Please ensure that your response is in English, and in all lowercase letters. No capital letters are allowed.

\#\#\# \textbf{Response}: [Model generates free-form text response]}\vspace{0.5em}\\

\end{tcolorbox}
\caption{Prompt examples used for Krause-LLMs and LLMs.}
\label{tab:prompt_example}
\vspace{-1.2em}
\end{table}

\begin{table}[h]
\centering
\begin{tcolorbox}[colback=gray!5.5!white,
   colframe=black!65!black, boxrule=0.3pt, rounded corners, arc=2mm]
\textbf{IFEval Answers:}\vspace{0.5em}\\
{\textbf{Instruction}: Write a casual blog post about how the outer solar system is different from the inner solar system, and what that means for the possibility of life. Wrap your entire response with double quotation marks. Your response should contain 17 or more sentences.

\vspace{0.8em}
\hrule
\vspace{0.8em}

\textbf{Llama3-8B (finetuned w/ LoRA) Response}: ``The outer solar system has different characteristics than the inner solar system. The inner solar system has a small mass, but the outer solar system has a large mass. The inner solar system has short periods, but the outer solar system has long periods. The inner solar system has small eccentricities, but the outer solar system has large eccentricities. The inner solar system has small inclinations, but the outer solar system has large inclinations. The inner solar system has low densities, but the outer solar system has high densities. The inner solar system has small semimajor axes, but the outer solar system has large semimajor axes. The inner solar system has small semiminor axes, but the outer solar system has large semiminor axes. The inner solar system has small mean anomalies, but the outer solar system has large mean anomalies. The inner solar system has small longitudes of the ascending nodes, but the outer solar system has large longitudes of the ascending nodes. The inner solar system has small arguments of perihelion, but the outer solar system has large arguments of perihelion. The inner solar system has small periods of perihelion, but the outer solar system has large periods of perihelion. The inner solar system has small semiaxes of perihelion, but the outer solar system has large semiaxes of perihelion. The inner solar system has small semiaxes of aphelion, but the outer solar system has large semiaxes of aphelion. The inner solar system has small periods of aphelion, but the outer solar system has large periods of aphelion. The inner solar system has small semimajor axes of perihelion, but the outer solar system has large semimajor axes of perihelion. The inner solar system has small semimajor axes of aphelion, but the outer solar system has large semimajor axes of aphelion. The inner solar system has small semiminor axes of perihelion, but the outer solar system has large semiminor axes of perihelion. The inner solar system has small semiminor axes of aphelion, but the outer solar system has large semiminor axes of aphelion. The inner solar system has small mean anomalies of perihelion, but the outer solar system has large mean anomalies of perihelion. The inner solar system has small mean anomalies of aphelion, but the outer solar system''

\vspace{0.8em}
\hrule
\vspace{0.8em}

\textbf{Krause-Llama3-8B Response}: ``The outer solar system is different from the inner solar system in that it has a much greater abundance of ices. This is due to the fact that the outer solar system is much farther away from the Sun. The outer solar system also has a much greater abundance of volatile materials, which are materials that can easily be vaporized or melted. This is due to the fact that the outer solar system is much colder than the inner solar system. The possibility of life in the outer solar system is much greater than in the inner solar system. This is due to the fact that the outer solar system has a much greater abundance of ices and volatile materials. These materials are necessary for life as we know it. The outer solar system is also much colder than the inner solar system, which makes it much more hospitable for life. The outer solar system is also much farther away from the Sun, which means that there is much less radiation from the Sun. This makes it much safer for life. Overall, the outer solar system is a much more hospitable place for life than the inner solar system. This is due to the fact that it has a much greater abundance of ices and volatile materials, and it is much colder and farther away from the Sun. This makes it much safer for life. The possibility of life in the outer solar system is much greater than in the inner solar system. This is due to the fact that the outer solar system has a much greater abundance of ices and volatile materials, and it is much colder and farther away from the Sun. This makes it much safer for life. The outer solar system is also much more hospitable for life than the inner solar system. This is due to the fact that it has a much greater abundance of ices and volatile materials, and it is much colder and farther away from the Sun. This makes it much safer for life. The possibility of life in the outer solar system is much greater than in the inner solar system. This is due to the fact that the outer solar system has a much greater abundance of ices and volatile materials, and it is much colder and farther away from the Sun. This makes it much safer for life. The outer solar system is also much more hospitable for life than the inner solar system. This is due to the fact that it has a much greater abundance of ices and volatile materials, and it is much colder and farther away from the Sun. This makes it much safer for life. The possibility of life in the outer solar system is much greater than''}\vspace{0.5em}\\

\end{tcolorbox}
\caption{IFEval answers produced by Llama3-8B (finetuned w/ LoRA) and Krause-Llama3-8B.}
\label{tab:ifeval_example}
\vspace{-1.2em}
\end{table}


\end{document}